\documentclass[journal]{IEEEtran}
\ifCLASSINFOpdf
\else
\fi
\usepackage{times}
\usepackage{epsfig}
\usepackage{graphicx}
\usepackage{amsmath}
\usepackage{dsfont}
\usepackage{amssymb}
\usepackage{url}
\usepackage{amsfonts}
\usepackage{subfigure}
\usepackage{multirow}
\usepackage{makecell}
\usepackage{booktabs}
\usepackage{lineno}
\usepackage{bbding}
\usepackage{pifont}
\usepackage{color}
\usepackage{xcolor}

\usepackage{lineno}
\begin{document}

\title{Prompting Video-Language Foundation Models with Domain-specific Fine-grained Heuristics for Video Question Answering}
\author{Ting~Yu,~\IEEEmembership{Member,~IEEE},
       Kunhao~Fu,
       Shuhui~Wang,~\IEEEmembership{Member,~IEEE},
       Qingming~Huang,~\IEEEmembership{Fellow,~IEEE},
       Jun~Yu,~\IEEEmembership{Senior Member,~IEEE}
    \thanks{This work was supported by National Natural Science Foundation of China under Grant No. 62002314 and Zhejiang Provincial Natural Science Foundation of China under Grant No. LY23F020005. }
    \thanks{T. Yu and K. Fu are with the School of Information Science and Technology, Hangzhou Normal University, Hangzhou 311121, China (e-mail: yut@hznu.edu.cn; fukunhao@stu.hznu.edu.cn).}
     \thanks{S. Wang is with the Key Laboratory of Intelligent Information Processing, Institute of Computing Technology, Chinese Academy of Sciences, Beijing 100190, China (e-mail: wangshuhui@ict.ac.cn).}
    \thanks{Q. Huang is with the School of Computer Science and Technology, University of Chinese Academy of Sciences, Beijing 101408, China (e-mail: qmhuang@ucas.ac.cn).}
   \thanks{J. Yu is with the Key Laboratory of Complex Systems Modeling and Simulation, School of Computer Science and Technology, Hangzhou Dianzi University, Hangzhou 310018, China (email: yujun@hdu.edu.cn).}
}
\maketitle

\begin{abstract}
Video Question Answering (VideoQA) represents a crucial intersection between video understanding and language processing, requiring both discriminative unimodal comprehension and sophisticated cross-modal interaction for accurate inference. Despite advancements in multi-modal pre-trained models and video-language foundation models, these systems often struggle with domain-specific VideoQA due to their generalized pre-training objectives. Addressing this gap necessitates bridging the divide between broad cross-modal knowledge and the specific inference demands of VideoQA tasks. To this end, we introduce HeurVidQA, a framework that leverages domain-specific entity-action heuristics to refine pre-trained video-language foundation models. Our approach treats these models as implicit knowledge engines, employing domain-specific entity-action prompters to direct the model's focus toward precise cues that enhance reasoning. By delivering fine-grained heuristics, we improve the model’s ability to identify and interpret key entities and actions, thereby enhancing its reasoning capabilities. Extensive evaluations across multiple VideoQA datasets demonstrate that our method significantly outperforms existing models, underscoring the importance of integrating domain-specific knowledge into video-language models for more accurate and context-aware VideoQA.
\end{abstract}


\section{Introduction}

\begin{figure}
\begin{center}
\includegraphics[width=0.5\textwidth]{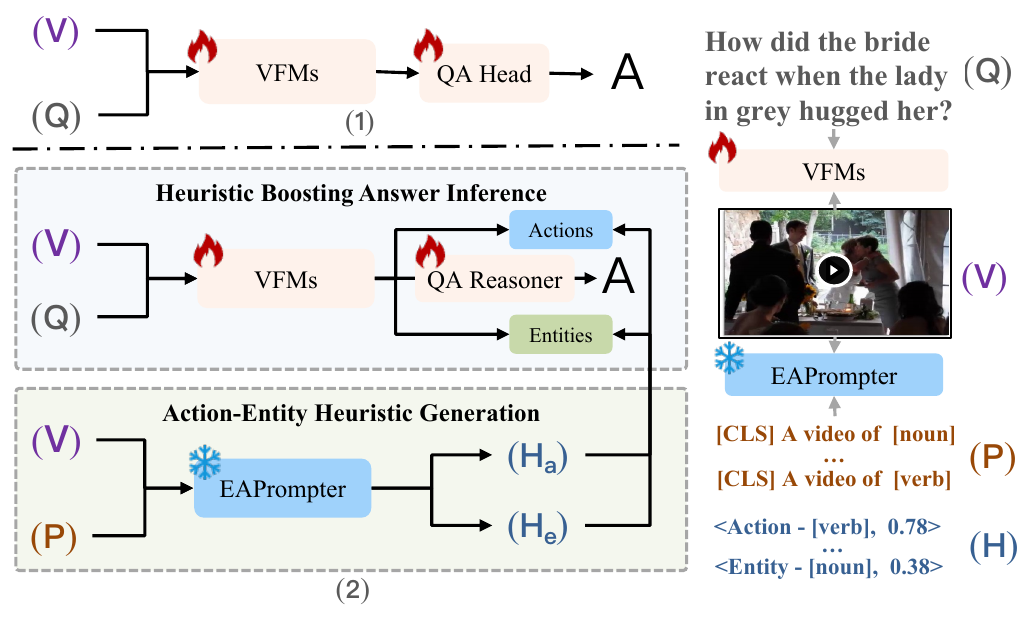}
\caption{Comparison of Recent Advanced VideoQA Models with the Proposed HeurVidQA Framework: (1) Recent methods utilize VFMs to generate general, domain-agnostic representations, complemented by task-specific heads for question-answering. (2) Our HeurVidQA framework enhances VFMs by integrating domain-specific entity-action prompts, blending implicit general knowledge with fine-grained domain-specific insights, to improve cross-modal understanding and reasoning in VideoQA tasks.
}
\label{fig:vqa_example}
\vspace{-10pt}
\end{center}
\end{figure}

\IEEEPARstart{T}{he} evolution of large-scale text pre-training has significantly advanced the capabilities of Large Language Models (LLMs) \cite{touvron2023llama, chung2022scaling, zhang2205opt} such as GPT-3 \cite{brown2020language} and BERT \cite{devlin2018bert}, empowering them with an unprecedented ability to comprehend and process complex linguistic structures. These breakthroughs have paved the way for sophisticated natural language processing applications \cite{peng2023you, zhu2022storytrans}. Building on the success of LLMs, the development of Large Multi-modality Pre-trained Models (LMMs) \cite{liu2024visual, li2023blip, lin2023video, achiam2023gpt}, exemplified by models like CLIP \cite{radford2021learning} and DALL-E \cite{ramesh2021zero}, represents a significant leap forward by integrating textual and visual data during pre-training. 
This fusion has driven remarkable progress in fields requiring deep cross-modal understanding, including robotics \cite{du2024learning}, medical diagnostics \cite{wang2023self, chiu2023toward}, and interactive gaming environments \cite{wang2023voyager}.

Video Question Answering (VideoQA) \cite{xiao2022video, zang2023discovering, gao2023mist, wang2023multi} stands out as a particularly challenging yet rapidly growing field within vision-language bridging research \cite{li2022blip, wang2023position}. It demands both discriminative unimodal understanding \cite{devlin2018bert, bertasius2021space, liu2023efficientvit} and comprehensive cross-modal interaction \cite{radford2021learning, li2022blip, li2021align, liu2023video}, integrating video understanding with language processing to infer reliable answers. While traditional approaches have explored enhanced video-linguistic models \cite{zhuang2020multichannel, zhang2020action} and adaptive cross-modal interactions \cite{chen2023tem}, the advent of Video Language Foundation Models (VFMs) \cite{li2022blip, wang2022omnivl, lei2021less}, such as ALPRO \cite{li2022align} and SiaSamRea \cite{yu2021learning}, has brought about a paradigm shift. These models, pre-trained on extensive video-text datasets \cite{xu2023youku, bain2021frozen}, have been pivotal in equipping machines with broad cross-modal knowledge, setting new standards for video content interpretation and advancing the boundaries of VideoQA \cite{xiao2022video, zang2023discovering, gao2023mist}.

Despite the advanced capabilities of VFMs, their generalized pre-training often underperforms in specialized VideoQA domains. This gap underscores the need for precise, context-aware adaptations to fully harness these models. Integrating prompt engineering with VFMs offers a promising solution by providing tailored prompts that steer the models toward domain-specific challenges, ensuring more accurate outcomes. However, the success of prompt-based approaches heavily relies on the design and quality of the prompts themselves. This is particularly true in VideoQA, where understanding the interplay between entities and their actions within videos is paramount. Generic prompts may not suffice for tasks that demand deep, context-specific insights, underscoring the need for a more refined strategy.

In this study, we introduce HeurVidQA, a novel framework designed to enhance video-language foundation models with domain-specific entity-action heuristics to address the complex demands of VideoQA. These complex demands arise from the need to process diverse and intricate video content, requiring not only an understanding of temporal sequences and spatial relationships but also the ability to handle varied question types that demand nuanced, context-aware reasoning. Our approach transforms these large-scale models into dynamic knowledge engines capable of navigating such intricate content with high precision and adaptability. We employ a strategy that integrates domain-specific, fine-grained heuristics into the prompt design, refining focus on essential cross-modal elements. 
As depicted in Figure \ref{fig:vqa_example}, we utilize instantiated prompt templates and specific regions within video frames to enable the prompter to accurately identify dynamic actions that vary spatially and evolve entities over time. This strategic use of heuristic-based prompts, informed by a deep understanding of key entities and their interactions, sharpens the model's focus and enhances analytical depth. Subsequently, leveraging these finely designed, context-aware entity-action heuristics, we prompt VFMs to generate coherent and contextually aligned responses.HeurVidQA underscores the criticality of recognizing key entities and actions for effective video content analysis. By ensuring the model's focus aligns with the question intricacies specific to a video, we enable a more contextually informed and impactful analysis, yielding answers that are accurate and deeply contextualized. 

In summary, our contributions are threefold:
\begin{itemize}
\item We introduce HeurVidQA, a framework that enhances video-language foundation models for VideoQA by integrating general implicit priors with domain-specific, fine-grained knowledge, thereby improving cross-modal understanding and reasoning capabilities.
\item We develop a domain-specific prompting mechanism utilizing fine-grained entity-action heuristics to guide the model in accurately identifying and interpreting dynamic actions and evolving entities within videos.
\item We validate the effectiveness of our approach through comprehensive evaluations on four VideoQA datasets across different domains, demonstrating robust and generalizable performance compared to existing methods.
\end{itemize}

\section{Related Work}\label{sec:related_work}
\subsection{Video Question Answering}
Video Question Answering (VideoQA) has emerged as a critical area in cross-modal research, driven by significant advancements in vision-language integration. Early efforts in this domain focused on adapting ImageQA models to video contexts, primarily leveraging LSTM-based encoders \cite{zeng2017leveraging, xu2017video}. 
However, these approaches were limited by their reliance on unimodal understanding, often failing to capture the inherent temporal dynamics of video content. This shortcoming underscores the necessity for advancing research towards cross-modal understanding approaches. To address this, attention mechanisms \cite{li2019beyond, jiang2020divide} have been investigated to extract linguistically guided visual features. Further advancements include the development of hierarchical attention mechanisms \cite{xu2017video, zhang2022erm}, co-memory networks \cite{gao2018motion}, and heterogeneous memory-enhanced models \cite{fan2019heterogeneous}, which have proven effective in modeling the complex interplay between video and question features.
The advent of the Transformer model \cite{vaswani2017attention} brought a revolutionary shift in natural language processing (NLP) and found application in VideoQA by Li \emph{et al.} \cite{li2019beyond}. This advancement significantly enhanced unimodal understanding within VideoQA. However, the visual modality’s limitations necessitated the development of novel cross-modal frameworks. Leveraging self-attention and co-attention mechanisms, Transformers facilitated deeper cross-modal interaction, enabling the extraction of pertinent visual features. Additionally, structured representations, including heterogeneous graph alignment networks \cite{jiang2020reasoning}, structure-aware interaction models \cite{park2021bridge}, and Dynamic Graph Transformers \cite{xiao2022video}, further advanced reasoning capabilities in VideoQA.
Recent studies in VideoQA have also tackled challenges related to confounders and compositional reasoning. Grounding indicators \cite{li2022invariant} have been developed to mitigate spurious correlations, while question decomposition engines \cite{gandhi2022measuring} offer valuable insights into compositional reasoning. To address confounding factors, multimodal causal reasoning frameworks \cite{zang2023discovering} have been introduced, along with adaptive spatial-temporal attention mechanisms \cite{gao2023mist}. As research into multi-granularity advances, emerging approaches increasingly explore the impact of varying granularities on VideoQA performance \cite{10130300,10547379,10092782,10508294}, focusing on multi-level intra- and inter-granularity relations to enhance cross-modal comprehension. Another approach \cite{li2022invariant, yu2023self} to solving VideoQA tasks focuses on identifying video frames relevant to the question, drawing inspiration from similar text-to-video retrieval (TVR) tasks. For example, Wu et al. \cite{wu2023empirical} explored both text-free and text-guided frame selection strategies in TVR, determining an optimal strategy that balances accuracy and computational efficiency.
In the context of multi-modal pre-training on large-scale vision-text data \cite{10187165, li2022blip, li2023blip, li2022align, li2021align, li2020oscar, radford2021learning}, Transformer-based models have demonstrated remarkable advancements. While prompt learning and engineering \cite{radford2021learning} have enhanced pre-training, the customization of prompts specifically tailored to entities and actions in videos remains underexplored. This paper addresses this gap by constructing task-specific prompts finely tuned to entities and actions, thereby enabling superior spatiotemporal reasoning in VideoQA. Our novel approach enriches video content understanding, pushing the boundaries of VideoQA performance. 
In this paper, we introduce a novel heuristic knowledge engine, EAPropmter, designed to extract detailed action and entity information embedded in videos. By leveraging diverse cropping strategies and customized prompt templates, EAPropmter significantly elevates the model’s ability to identify and interpret fine-grained information. This innovative data-centric approach offers a groundbreaking solution to VideoQA, paving the way for further exploration in this field.

\subsection{Vision-and-Language Pre-training}
The paradigm of pre-training on large-scale visual-language data followed by fine-tuning for specific downstream tasks has demonstrated significant success in cross-modal applications, including VideoQA. Vision-and-Language Pre-trained (VLP) models \cite{radford2021learning, jia2021scaling}, such as CLIP \cite{radford2021learning} and ALIGN \cite{jia2021scaling}, commonly use standardized pre-training objectives like masked language modeling (MLM) and image-text matching (ITM) on extensive vision-text datasets. However, earlier VLP models, such as OSCAR \cite{li2020oscar}, depended heavily on pre-trained object detectors for visual feature extraction. This reliance led to increased computational overhead and introduced noise into the image-text data, which ultimately impacted downstream performance.
Despite their strong performance across various tasks, these VLP models face two significant limitations: first, they often struggle to effectively model complex vision-text interactions, and second, they tend to overfit to the noisy data prevalent in large-scale image-text datasets, which ultimately hampers their generalization performance.
To overcome these limitations, various strategies have been proposed. For instance, Li \emph{et al.} \cite{li2021align} introduced an intermediate image-text contrastive loss to enhance cross-modal alignment and semantic understanding while mitigating the effects of noisy data using Momentum Distillation. Similarly, BLIP \cite{li2022blip} advanced the quality of vision-text pairs through the Captioning and Filtering (CapFilt) approach, which fine-tunes a captioner using language modeling and a filter with cross-modal tasks, thereby improving the model's handling of vision-text relationships.
In the domain of video processing, VideoBERT \cite{sun2019videobert} extended the BERT model to video analysis but neglected the crucial roles of textual cues and cross-modal interactions. Meanwhile, ActBERT \cite{zhu2020actbert} depended on a pre-trained object detector to generate object pseudo-labels, but its limited detection categories and high computational costs resulted in suboptimal performance. Moving beyond one-directional transfer models, unified transformer-based architectures, as proposed by Wang \emph{et al.} \cite{wang2022omnivl}, have emerged to handle multimodal input sources, enabling joint pre-training for image-language and video-language tasks, thus benefiting both image and video-related assignments.
More recently, with the advancement of prompt learning, VLP models have benefited from the integration of carefully designed prompts, which have significantly enhanced their performance in downstream tasks. This approach allows for greater control and precision, enabling models to produce task-specific outcomes more effectively. The success of prompt learning lies in its ability to address challenges such as limited data and complex annotations, positioning it as a promising direction for advancing cross-modal research. By incorporating these techniques, researchers have achieved impressive results, further narrowing the gap between vision and language understanding.
Our proposed HeurVidQA framework marks a significant departure by incorporating novel visual heuristics that enhance the VLP model with fine-grained perceptual capabilities. Leveraging the robust cross-modal learning strengths of these models, our approach not only improves their ability to capture intricate visual nuances but also bridges the gap between pre-training and downstream task execution. This integration of new visual cues within HeurVidQA represents a pivotal advancement in refining and advancing state-of-the-art visual language processing.

\subsection{Prompt Learning}
\begin{figure*}
\begin{center}
\includegraphics[width=1\textwidth]{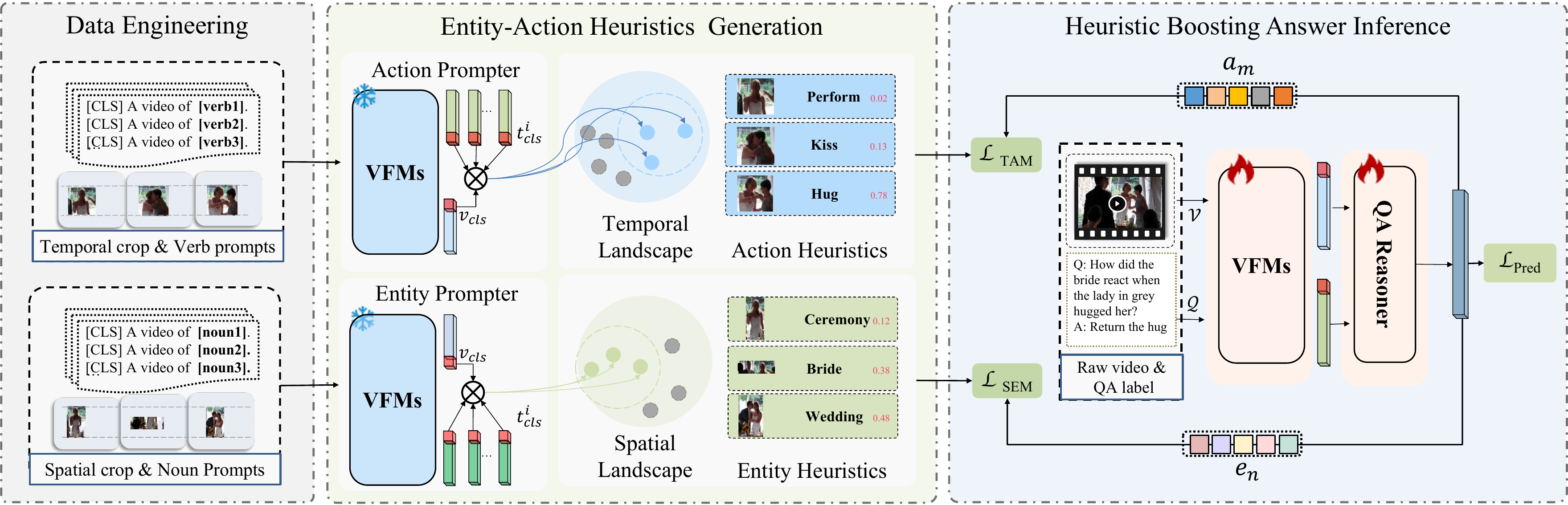} 
\caption{Overview of the Proposed HeurVidQA Framework. The framework consists of two primary stages: Entity-Action Heuristic Generation and Heuristic-Enhanced Answer Inference. In the first stage, EAPrompter extracts action heuristics through temporal-aware action detection and entity heuristics via spatial-aware entity detection using spatial-temporal-aware crops and instantiated prompt templates. These heuristics, sourced from a predefined vocabulary, include confidence scores for each detected action and entity. The second stage utilizes a Vision-Language Foundation Model (VFM) to process multimodal features through self-attention and cross-attention modules, resulting in a semantically enriched cross-modal fusion that supports both final answer prediction and anticipation of actions and entities in the video.}
\label{fig:overview}
\end{center}
\vspace{-15pt}
\end{figure*}

Prompt learning \cite{brown2020language, li2021prefix, petroni2019language} has emerged as a powerful technique in the cross-modal field, allowing models to achieve task-specific results by incorporating targeted prompts during training. This approach enhances control and accuracy, providing a promising solution to challenges posed by limited data and complex annotations.
At its core, prompt learning can be likened to a "fill-in-the-blank" task for pre-training models. By masking certain words in sentences and prompting the model to predict the missing tokens, the model learns to comprehend contextual relationships between words in a sentence. CLIP \cite{radford2021learning}, in particular, showcases the efficacy of prompt-based learning for image recognition. Through contrastive learning on a large-scale noisy dataset, it aligns relevant image-text pairs and distinguishes unrelated ones, effectively achieving image-text alignment. For image recognition, it incorporates the label into a descriptive sentence, enabling the model to predict specific words (e.g., ``An image about a [MASK]"). Remarkably, CLIP achieves impressive accuracy in image recognition tasks even without fine-tuning, thanks to the flexibility of prompt design, making it applicable to various image-text tasks.
Building on the success of CLIP, ALIGN \cite{jia2021scaling} further advances vision-language models by training on a massive dataset comprising 1.8 billion image-text pairs. ALIGN surpasses CLIP’s predictive performance, achieving superior results. Meanwhile, CoOp \cite{zhou2022learning} introduces a prompt fine-tuning method, adapting NLP techniques to the cross-modal domain. By employing learnable tokens as prompts and minimizing classification losses, CoOp effectively learns Soft Prompts with minimal annotated data, overcoming the limitations of manually tuned prompts. This innovation yields significant performance improvements and greater flexibility.
However, CoOp’s learned context struggles to generalize to unseen classes, a challenge addressed by CoCoOp \cite{zhou2022conditional}. CoCoOp generates input-conditional vectors via a lightweight network, dynamically adjusting prompts to improve generalization across varying classes. This method enhances performance on unseen classes, solidifying CoCoOp’s robustness.
Additionally, Ma \emph{et al.} \cite{10045664} tackles the overfitting issue in CoOp by projecting gradients onto a low-rank subspace during back-propagation.
Exploring prompt techniques further, researchers have focused on enhancing tasks through verb and noun prompts. For instance, Li \emph{et al.} \cite{10360871} introduce CSA-Net, leveraging the RegionCLIP \cite{zhong2022regionclip} model to create fine-grained regional semantic visual spaces, which enhances content comprehension. While prompt learning has been predominantly explored in image-based tasks, Li \emph{et al.} \cite{li2022align} extend these innovations to video models, using a pre-trained entity prompter to generate pseudo-labels, thereby achieving notable improvements across various benchmarks.
However, video content goes beyond just entities; it crucially involves actions that convey vital information. Recognizing this, our approach leverages both entities and actions in the video to construct heuristics, empowering our model with enhanced temporal and spatial reasoning capabilities for video-based question answering. Incorporating these fine-grained details has led to state-of-the-art performance on the challenging NExT-QA \cite{xiao2021next} dataset, demonstrating the effectiveness of our approach in advancing video question answering tasks within the cross-modal domain.

\section{Method}\label{sec:mfb}
The overall architecture of our HeurVidQA is illustrated in Figure \ref{fig:overview}.
In our approach, we seek to accurately predict the answer $y$ to a given question $\mathcal{Q}$, posed in natural language, based on the content of a raw video $\mathcal{V}$. The following equation formalizes this objective:
\begin{equation}
\hat{y} = \underset{y \in \mathcal{A}} {\operatorname{argmax}}\mathcal{F}_W (y|\mathcal{V}, \mathcal{Q}, \mathcal{A}),
\end{equation}
where $\mathcal{F}_W$ represents the VideoQA model parameterized by weights $W$. 
The function's output, $\hat{y}$, varies based on the VideoQA task configuration, offering an open-ended response from a global answer set or a selected choice from multiple alternatives. 
For multiple-choice QA, we establish a linkage between the question and candidate answers with a [PAD] token, supplemented by the inclusion of a specialized [CLS] token positioned at the inception of the textual sequence:

\begin{equation}
  \mathcal{T}^{MC}(Q, a_i|a_i\in A_{MC})=\mathrm{[CLS]} \  Q \ \mathrm{[PAD]} \ a_i,
\end{equation}
where $\mathcal{T}^{MC}$ is the conversion function that converts multiple-choise question $Q$ and its candidate answers $a_i$ to the input format. For open-ended QA, where no predefined candidate answers exist, the [CLS] token is directly associated with the commencement of the question. The formulation of the conversion function is as follows:
\begin{equation}
  \mathcal{T}^{OE}(Q)=\mathrm{[CLS]} \ Q.
\end{equation}
Our model is designed to be adaptable across these varied formats, enabling comprehensive applicability within the field of VideoQA.

In the following, we outline the workflow of the proposed HeurVidQA framework, organized into three key phases. The \textbf{Data Engineering} phase establishes the foundation by optimizing visual inputs for the EAPrompter through strategic clipping and crafting video entity-action templates. Next, the \textbf{Entity-Action Heuristics Generation} subsection introduces the EAPrompter’s architecture and training, focusing on synthesizing heuristic insights from the processed data. Finally, the \textbf{Heuristic Boosting Answer Inference} phase utilizes these heuristics to enable synergistic interaction between VFMs and the knowledge engine, enhancing answer inference performance.

\subsection{Data Engineering}
Our HeurVidQA framework begins by preparing videos and prompts for analysis. We segment the videos temporally and spatially, targeting key moments and areas likely to hold relevant information. Simultaneously, we generate domain-specific prompts using action-related verbs and object-focused nouns anticipated in the video content. These tailored prompts guide the foundation models in identifying critical elements. This phase is crucial for data preparation, ensuring the subsequent heuristic generation stages are well-informed and effective.
To enhance both temporal and spatial sensitivity in video cropping, we employ distinct strategies tailored to each dimension. Temporal sensitivity is addressed by consistently cropping identical spatial regions across keyframes, ensuring focused capture of evolving actions. For spatial sensitivity, we extract varied regions across different frames, providing a comprehensive spatial understanding of the entities involved. The resulting video segments are denoted as $\hat{V^t}$ for temporally cropped segments and $\hat{V^s}$ for spatially cropped segments.
Additionally, the textual component is refined through a prompt engineering process. We employ pre-defined prompt templates (detailed in Table \ref{table:prompt-templates}) for action and entity identification, such as ``A video of [NOUN]" and `` A video of a [VERB]." These placeholders are systematically replaced with verbs and nouns extracted from the top 1,000 most frequent terms found in domain-specific QA pairs. This approach ensures that the prompts are highly relevant to the domain, enhancing the model's ability to generate precise and context-aware responses.
This preprocessing stage primes our model by equipping it with fine-grained visual cues and aligning it with contextually relevant textual prompts, laying a robust foundation for the extraction of meaningful entity-action heuristics.

\subsection{Entity-Action Heuristics Generation}
After data engineering, our HeurVidQA framework progresses to the crucial stage of generating domain-specific fine-grained heuristics. This phase employs a two-pronged approach: (1) Action Heuristics Generation focuses on capturing verbs and their temporal dynamics within video content, enabling the model to comprehend ongoing activities; (2) Entity Heuristics Generation targets the identification and tracking of nouns and their spatial attributes, ensuring the accurate reference of key subjects within the visual space. This dual approach ensures that both temporal and spatial aspects are effectively captured, enhancing the model’s capacity for precise video question answering.

\subsubsection{EntityActionPrompter}\label{sec:eap}
\begin{figure}
\begin{center}
\includegraphics[width=0.5\textwidth]{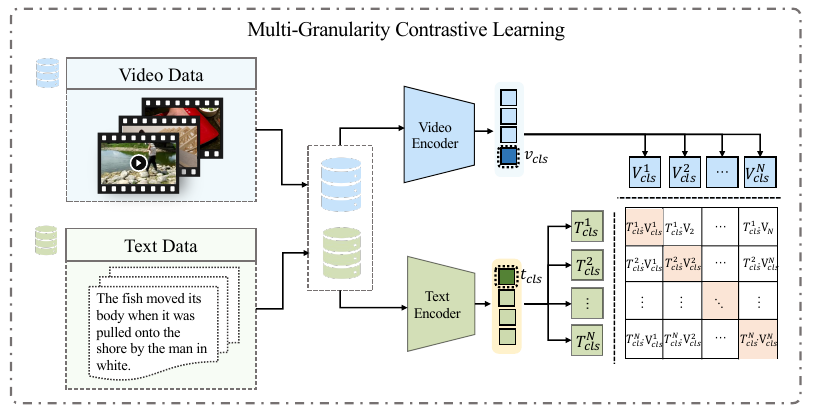}
\caption{Training Framework of EAPrompter. The Symmetric Contrastive Loss is employed to jointly train a video encoder and a text encoder, facilitating the prediction of correct pairings within batches of \{video, text\} samples.}
\label{fig:eaprompter}
\vspace{-20pt}
\end{center}
\end{figure}
The core of our heuristic generation lies within the EntityActionPrompter \emph{(EAPrompter)}, which operates through two sub-modules: the action prompter \emph{(AP)} and the entity prompter \emph{(EP)}, each tailored to capture distinct yet complementary aspects of the video content's entity-action landscape. The \emph{AP} enhances temporal precision by randomly cropping identical spatial regions across sparsely selected video frames, while the \emph{EP} augments spatial acuity by extracting varied regions across different frames. Both modules share an identical architecture, adopting standard clip-based visual-language model structures with a video branch and a text branch. For the video branch, both \emph{AP} and \emph{EP} employ a 12-layer TimeSformer$_{224}$ \cite{bertasius2021space} to capture video embeddings. It takes as input a clip $V \in \mathbb{R}^{C \times F \times H \times W }$, which undergoes random sampling, scaling, cropping, and random masking to enhance computational efficiency and consistent performance.

To facilitate the formula expression, the unimodal visual embeddings derived from \emph{AP} are denoted as $\{v^t_{cls}, v^t_1, \dots, v^t_{N^t_v}\}$, where $v^t_{cls}$ and $N^t_v$ corresponds to video global temporal embedding and temporal cropping numbers respectively. Visual embeddings from \emph{EP} are denoted as $\{v^s_{cls}, v^s_1, \dots, v^s_{N^s_v}\}$, where $v^s_{cls}$ and $N^s_v$ are video global spatial embedding and spatial cropping numbers. 
For the text branch, we employ a multi-layer bidirectional transformer \cite{devlin2018bert} to encode text semantics in a hierarchical collaborative parsing pattern. This procedure produces corresponding prompts embeddings, $\{t^{a_1}_{cls}, t_{cls}^{a_2}, \dots,t_{cls}^{a_M}\}$ and $\{t^{e_1}_{cls}, t_{cls}^{e_2}, \dots,t_{cls}^{e_N}\}$ for action and entity prompts, respectively.
For cross-modality alignment, a similarity function $s(\cdot)$ is optimized between the video and textual embeddings as follows:
\begin{equation}
  s(v_{cls}, t_{cls})=f_v(v_{cls}) \cdot f_t(t_{cls}),
\end{equation}
where a higher similarity score is obtained for matching videos and prompts. 

Both \emph{AP} and \emph{EP} are pre-trained on webly sourced video-text pairs with video-text contrastive (VTC) loss. 
As depicted in Figure \ref{fig:eaprompter}, the symmetric contrastive loss plays a key role in enhancing cross-modal alignment. By pulling positive pairs closer (on the diagonal) and pushing negative pairs further apart (off-diagonal), this mechanism drives the encoder representations to converge in a low-dimensional space, thereby optimizing inter-modal alignment. The symmetric contrastive loss for the vision-to-text modality is calculated as follows:

\begin{equation}
  \mathcal{L}_{v2t}=-\sum_{i=1}^{B}{\log{\frac{exp(s(v_{cls}^i, t_{cls}^i)/\tau)}{\sum^B_{j=1}{exp(s(v_{cls}^i, t_{cls}^j)/\tau)}\ }}} .
\end{equation}
The symmetric contrastive loss for the text-to-vison contrastive loss is computed as:
\begin{equation}
  \mathcal{L}_{t2v}=-\sum_{i=1}^{B}{\log{\frac{exp(s(t_{cls}^i, v_{cls}^i)/\tau)}{\sum^B_{j=1}{exp(s(t_{cls}^i, v_{cls}^j,)/\tau)}\ }}},
\end{equation}
where $ B$ represents the batch size and $\tau$ signifies a learnable temperature parameter, the prompter, after being pre-trained on extensive datasets, exhibits strong alignment capabilities for video and textual data. To prevent external noise from affecting its performance, we freeze the prompter's parameters during the entity-action heuristic generation stage. Through this approach, the EAPrompter effectively constructs a detailed entity-action map of the video content, providing a heuristic-driven analysis that greatly enhances the accuracy and relevance of the subsequent answer inference stage.

\subsubsection{Heuristic Generation}
\begin{figure}
\begin{center}
\includegraphics[width=0.4\textwidth]{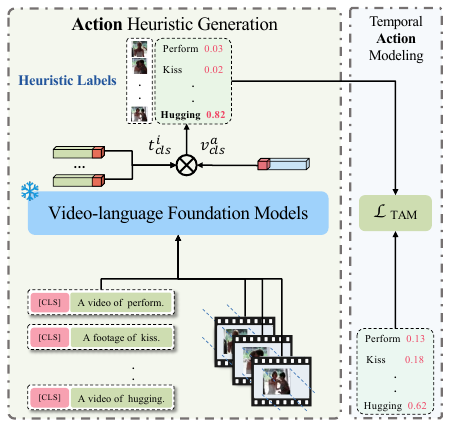}
\caption{Operational Principle of Action Prompter. The Action Prompter processes instantiated templates alongside sparsely selected video frames, which have been refined through a space-time cropping strategy.}
\label{fig:aprompter}
\vspace{-20pt}
\end{center}
\end{figure}

\begin{figure*}
\begin{center}
\includegraphics[width=1\textwidth]{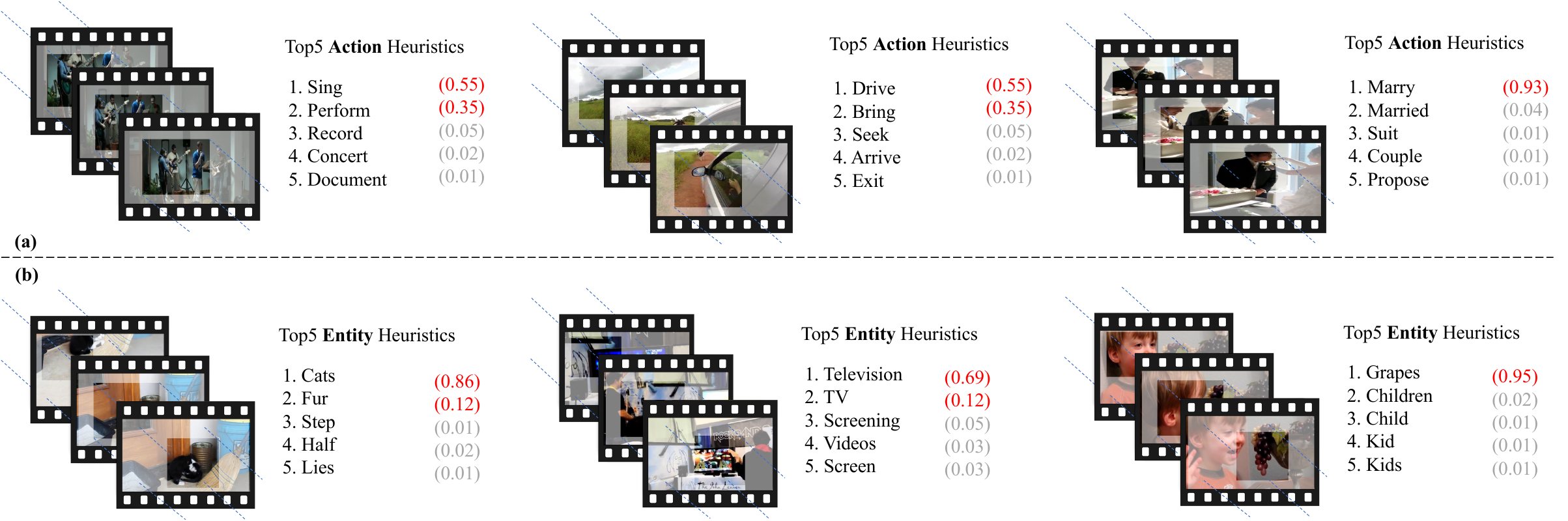}
\caption{
Visualizations of heuristics generated by EAPrompter. We selected commonly used verbs and nouns as candidates for action and entity prompts, respectively, and labeled the top five with their associated probabilities (in brackets). To obtain heuristics for actions \textbf{(a)} and entities \textbf{(b)}, we applied distinct processing strategies: consistent regions across frames for action prompts, and varying regions for entity prompts to mimic real-world scenarios. Labels with a probability below 10\% were filtered out, with significant heuristics highlighted in red.
}
\label{fig:pseudo_labels}
\end{center}
\end{figure*}

In the Heuristic Generation subsection, we introduce a key innovation in our approach: the utilization of latent heuristic information embedded within video frames for VideoQA tasks. This involves generating heuristic labels through the action prompter module, as depicted in Figure \ref{fig:aprompter}. The action prompter processes instantiated templates alongside carefully selected video frames, applying a precise space-time cropping strategy to capture rich action information. By leveraging the knowledge gained during pre-training, the prompter aligns these templates with the corresponding frames, resulting in heuristic labels characterized by detailed probability distributions. Specifically, the action heuristics can be derived by calculating the softmax-normalized similarity between the temporal embeddings of the video, denoted as $\hat{v}^t_{cls}$, and the action prompt embeddings, represented as ${t_{cls}^{a_m}}^M_{m=1}$. This formulation is mathematically expressed as follows:
\begin{equation}
  h_{\hat{v}^t, a_m}= \frac{exp(s(\hat{v}^t_{cls}, t^{a_m}_{cls})/\tau)}{\sum^M_{m=1}exp(s(\hat{v}_{cls}^t, t^{a_m}_{cls})/\tau)\ }.
\end{equation}
The entity heuristics, produced by calculating the softmax-normalized similarity between the spatial embeddings of the video $\hat{v}^s_{cls}$ and all the entity prompt embeddings $\{t_{cls}^{e_n}\}^N_{n=1}$, can be formulated as follows:
\begin{equation}
  h_{\hat{v}^s, e_n} = \frac{exp(s(\hat{v}^s_{cls}, t^{e_n}_{cls})/\tau)}{\sum^N_{n=1}exp(s(\hat{v}_{cls}^s, t^{e_n}_{cls})/\tau)\ }.
\end{equation}
Figure \ref{fig:pseudo_labels} illustrates several examples of the entity-action heuristics generated by our approach. Our approach differs from traditional object detection methods that require substantial computing power for object identification. our method leverages targeted heuristic generation, by efficiently extracting relevant entity and action information across temporal and spatial dimensions, reducing the computational burden while maintaining high precision in VideoQA tasks.

\subsection{Heuristic Boosting Answer Inference}\label{sec:predictor}
Heuristic boosting answer inference stage leverages heuristic labels from the previous phase to strengthen interactions between VFMs and the heuristic knowledge engine, refining the model’s ability to infer answers accurately.
HeurVidQA utilizes VFMs to capture contextual video-language embeddings, followed by a QA reasoner to conduct question-answering interaction and communication. The QA reasoner stacks cross-modality alignment layers to fuse visual-textual embeddings to explore in-depth, informative clues. Specifically, it comprises several self-attention blocks, cross-attention layers, and feed-forward networks to deliver cross-modal fusions $\{e_{cls}, e_1, \dots, e_{N_v+N_t}\}$, where $e_{cls}$ serves as a pivotal link between the EAPrompter and the QA Reasoner. 
Instead of directly utilizing the fused embedding $e_{cls}$ to predict the answer,  HeurVidQA skillfully introduces the generated domain-specific entity-action heuristics as supervisory targets with Temporal Action Modeling (TAM) and {Spatial Entity Modeling (SEM)} losses. TAM Loss enhances the model's ability to understand temporal action sequences, while SEM Loss improves the recognition of spatial relationships among entities. These loss functions are pivotal in transferring the latent knowledge from the prompters to the baseline model, guiding the VFMs toward more precise reasoning in VideoQA tasks.
Through two distinct classifiers $c_1(\cdot)$ and $c_2(\cdot)$, we initially obtained the normalized confidence scores of action and entity heuristics:

\begin{equation}
  p_{\hat{a}, a_m}=c_1(e_{cls}), \qquad
  p_{\hat{e}, e_n}=c_2(e_{cls}),
\end{equation}
where $p_{\hat{a}, a_m}$ and $p_{\hat{e}, e_n}$ denote the classifier-derived probability distributions for action and entity heuristics, respectively. The subscripts $a_m$ and $e_n$ correspond to the $m^{th}$ action and $n^{th}$ entity within the respective distributions.

The TAM loss is calculated as the cross-entropy between the classifier-derived action heuristic distribution $p_{\hat{a}}$ and the target distribution $h_{\hat{v}_{cls}, a_m}$:
\begin{equation}
  \mathcal{L}_{TAM} = -\sum_{m=1}^M{h_{\hat{v}_{cls}, a_m} \cdot \log{p_{\hat{a}, a_m}}}.
\end{equation}
Similarly, the SEM loss is calculated as the cross-entropy between the classifier-derived entity heuristic distribution $p_{\hat{e}}$ and the target distribution $h_{\hat{v}_{cls}, e_n}$:
\begin{equation}
  \mathcal{L}_{SEM} = -\sum_{n=1}^N{h_{\hat{v}_{cls}, e_n} \cdot \log{p_{\hat{e}, e_n}}}.
\end{equation}
Guided by the supervision of entity-action heuristics, the fusion embedding retrieves the unimodal information that might have been lost during cross-modal interaction, enabling HeurVidQA to achieve fine-grained spatiotemporal reasoning. To predict the final answer $\hat{y}$, we use an independent classifier, which takes the heuristic-enhanced fusion representation as input, formulated as follows:
\begin{equation}
  \hat{y} = c_3(e_{cls}).
\end{equation}
Comparing the predictive answers $\hat{y}$  with the ground truth $y$,  we calculate the prediction loss as follows: 
\begin{equation}
  \mathcal{L}_{pred} = -\sum y\log{\hat{y}}.
\end{equation}
The complete training objective of HeurVidQA can be obtained by combining the prediction loss with the heuristic losses as follows: 
\begin{equation}
  \mathcal{L} = \mathcal{L}_{pred} + \alpha \mathcal{L}_{TAM} + (1-\alpha)\mathcal{L}_{SEM}.
\end{equation}
Where $\alpha$ represents a hyperparameter set to 0.5 in our experiments, it serves to balance the weighting between action and entity prompts. To explore other configurations, we introduced a dynamic gating mechanism that adjusts the influence of action and entity prompts based on the input data and training stage. This question-guided gating mechanism involves the following steps: processing the text input via VFMs to extract the global $t_{cls}$ embedding and passing this embedding through a gating network composed of two multi-layer perceptrons, a dropout function, and a sigmoid activation function.
\begin{equation}
  g(t_{cls}) = \mathrm{Sigmoid(MLP_2(Dropout(MLP_1}(t_{cls})))).
\end{equation}
Through the gating mechanism, we obtain a scalar value ranging from 0 to 1, representing the gate weight for the action prompt. Next, the gate weight for the entity prompt is calculated by subtracting the action prompt's weight from 1. Finally, these gate weights are applied to the TAM loss and SEM loss. The combined loss function is then formulated as follows:
\begin{equation}
  \mathcal{L} = \mathcal{L}_{pred} + g \mathcal{L}_{TAM} + (1-g)\mathcal{L}_{SEM}.
\end{equation}
This adaptive gating scheme has demonstrated its effectiveness in enhancing both the model's adaptability and stability by dynamically adjusting the influence of action and entity prompts.

\section{Experiments}\label{sec:experiments}
 We have conducted an extensive analysis to evaluate the performance of HeurVidQA across several established VideoQA benchmark datasets, including NExT-QA \cite{xiao2021next}, MSVD-QA \cite{xu2017video}, MSRVTT-QA \cite{xu2017video}, and SUTD-TrafficQA \cite{xu2021sutd}.

\subsection{Datasets}
\begin{itemize}
    \item \textbf{NExT-QA} \cite{xiao2021next} presents complex challenges requiring advanced causal and temporal reasoning. It comprises 5,440 videos averaging 44 seconds in length, with approximately 52K annotated QA pairs categorized into \emph{causal}, \emph{temporal}, and \emph{descriptive} questions. We also consider the $ATP_{hard}$ split from Buch \emph{et al.} \cite{buch2022revisiting}, featuring questions that demand sophisticated video comprehension beyond single-frame analysis.
 \item \textbf{MSVD-QA} \cite{xu2017video} is derived from the MSVD dataset and includes 10s video clips across 1,970 videos. It features around 51K QA pairs automatically generated based on video captions, covering diverse topics like video content, object recognition, and scene understanding.
 \item \textbf{MSRVTT-QA} \cite{xu2017video} is similar to MSVD-QA but on a larger scale, with 10K videos averaging 15 seconds each and containing 244K QA pairs. These are generated from video descriptions and serve to train and evaluate models in open-ended question answering.
\item \textbf{SUTD-TrafficQA} \cite{xu2021sutd} serves as a critical resource for advancing research in traffic-related question answering, requiring a deep understanding of traffic events and their causal relationships. It comprises over 10,000 videos depicting diverse traffic incidents, supplemented by 62,535 human-annotated QA pairs—56,460 for training and 6,075 for testing. This comprehensive dataset rigorously evaluates the cognitive abilities of models in understanding complex traffic scenarios. All tasks follow a multiple-choice format without constraints on the number of possible answers.
\end{itemize}

\subsection{Implementation Details}
Our method builds on ALPRO \cite{li2022align} as the video-language foundation model, fine-tuned using 4 NVIDIA GeForce 3090 GPUs. We used weights pre-trained on WebVid-2M and CC-3M datasets, alongside prompter weights pre-trained on noisy video-text pairs. The AdamW optimizer was employed, with a weight decay of 0.001 and a unified learning rate of 5e-5 across datasets. A linear decay schedule was used to adjust the learning rate dynamically, promoting rapid convergence initially and precise convergence during later training stages.
For raw video processing, videos were rescaled to $224 \times 224$, and a random sparse sampling strategy was applied to extract 16 frames while preserving their sequential order. The EAPrompter followed a distinct processing strategy, resizing videos to $256 \times 256$ before cropping a $224 \times 224$ area. After random sampling, 50-70\% of the original spatial area was randomly masked to capture specific actions and entities effectively. For the Entity-Action Heuristic Generation, spaCy was used to extract the top 1,000 frequent verbs and nouns as action and entity candidates. Heuristics were discarded if the highest-scoring entity had a normalized similarity score below 0.1. The training was conducted over 10 epochs for NExT-QA and SUTD-TrafficQA, and 15 epochs for MSVD-QA and MSRVTT-QA datasets.

\begin{table*} 
\centering
\caption{Comparison with state-of-the-art results on the validation standard and test sets of the NExT-QA dataset. Acc@C, Acc@T, and Acc@D denote accuracy for Causal, Temporal, and Descriptive questions, respectively, expressed as a percentage (\%). The \textbf{best} result is highlighted in bold, and the \underline{2nd best} result is underlined.
}
\label{table:test-std}
\begin{tabular}{lccccccccc}
\toprule
\multirow{2}*{\textbf{Method}} & \multirow{2}*{\textbf{Pre-trained}} &  \multicolumn{4}{c}{{\textbf{NExT-QA Val}}}  & \multicolumn{4}{c}{{\textbf{NExT-QA Test}}} \\
\cline{3-10}
\rule{0pt}{9pt}\makecell[c]  &  & \textbf{Acc@C} & \textbf{Acc@T} & \textbf{Acc@D} & \textbf{Acc@All} & \textbf{Acc@C} & \textbf{Acc@T} & \textbf{Acc@D} & \textbf{Acc@All}\\
\midrule
{\textcolor{gray}{SeViLA (2023) \cite{yu2023self}}} & - & \textcolor{gray}{73.4} & \textcolor{gray}{68.8} & \textcolor{gray}{83.5} &  \textcolor{gray}{73.4} & \textcolor{gray}{-} & \textcolor{gray}{-} & \textcolor{gray}{-} & \textcolor{gray}{-}\\
{\textcolor{gray}{LSTP (2024) \cite{wang2024lstp}}} & - & \textcolor{gray}{72.8} & \textcolor{gray}{66.5} & \textcolor{gray}{81.2} &  \textcolor{gray}{72.1} & \textcolor{gray}{-} & \textcolor{gray}{-} & \textcolor{gray}{-} & \textcolor{gray}{-}\\
{\textcolor{gray}{CREMA (2024) \cite{yu2024crema}}} & - & \textcolor{gray}{74.4} & \textcolor{gray}{69.4} & \textcolor{gray}{81.6} &  \textcolor{gray}{73.9} & \textcolor{gray}{-} & \textcolor{gray}{-} & \textcolor{gray}{-} & \textcolor{gray}{-}\\
\midrule
{CoMem (2018) \cite{gao2018motion}} & \ding{55} & 45.2 & 49.1 & 55.3 & 48.0 & 45.9 & 50.0 & 54.4 &  48.5\\
{EVQA (2019) \cite{zeng2017leveraging}} & \ding{55} & 42.5 & 46.3 & 45.8 &  44.2 & 43.3 & 46.9 & 45.6 & 44.9\\
{HME (2019) \cite{fan2019heterogeneous}} & \ding{55} & 46.2 & 48.2 & 58.3 &  48.7 & 46.8 & 48.9 & 57.4 & 49.2\\
{HCRN (2020) \cite{le2020hierarchical}} & \ding{55} & 45.9 & 49.3 & 53.7 &  48.2 & 47.0 & 49.3 & 54.0 & 48.9\\
{HGA (2020) \cite{jiang2020reasoning}} & \ding{55} & 46.3 & 50.7 & 59.3 &  49.7 & 48.1 & 49.1 & 57.8 & 50.0\\
{IGV (2022) \cite{li2022invariant}} & \ding{55} & - & - & - &  - & 48.6 & 51.7 & 59.6 & 51.3\\
{VGT (2022) \cite{xiao2022video}} & \ding{55} & 52.3 & 55.1 & 64.1 &  55.0 & 51.6 & 51.9 & 63.7 &  53.7\\
{CoVGT (2023) \cite{xiao2023contrastive}} & \ding{55} & \underline{58.8} &  57.4 &  69.4 &  60.0 & \textbf{58.5} & 57.0 & 66.8 &  59.4\\
\midrule
{ALPRO (2022) \cite{li2022align}} & \ding{51}  & 57.6 & 57.8 & 71.7 & 59.9 & 55.8 & 56.0 & \textbf{71.9} &  58.3\\
{VGT (PT) (2022) \cite{xiao2022video}} & \ding{51} & 53.4 & 56.4 & 69.5 &  56.9 & 52.8 & 54.5 & 67.3 &  55.7\\
{MIST-CLIP (2023) \cite{gao2023mist}} & \ding{51}  & 54.6 & 56.6 & 66.9 & 57.2 & - & - & - &  - \\
{$\mathrm{SeViT_{FiD}}$ (2023) \cite{kim2023semi}} & \ding{51}  & {54.0} & {54.1} & {71.3} &{56.7} & {-} & {-} & {-} &{55.2}\\
{VFC (2023) \cite{momeni2023verbs}} & \ding{51}  & 57.6 & 53.3 & 72.8 & 58.6 & - & - & - & -\\
{CoVGT (PT) (2023) \cite{xiao2023contrastive}} & \ding{51} & \textbf{59.7} &  58.0 &  69.9 &  60.7 & 58.0 & \textbf{58.0} & 68.4 &  \underline{59.7}\\
\midrule
{HeurVidQA (Ours)} & \ding{51}  & \underline{58.8} & \textbf{58.4} & \textbf{73.0} & \textbf{60.9} & \underline{58.1} & \underline{57.1} & \underline{71.1} & \textbf{59.9}\\
\bottomrule
\label{table:compare_1}
\end{tabular}
\end{table*}

\subsection{Experimental Results}
We evaluated our method against several state-of-the-art (SoTA) models on four VideoQA datasets: NExT-QA, SUTD-TrafficQA, MSVD-QA, and MSRVTT-QA. Our comparisons included non-pretrained models like VGT \cite{xiao2022video} and CoVGT \cite{xiao2023contrastive}, as well as pre-trained VL models, such as MIST \cite{gao2023mist} and our baseline, ALPRO \cite{li2022align}. 
\begin{table*}
\centering
\caption{Comparative experimental results of the proposed HeurVidQA model and existing state-of-the-art (SOTA) models on the MSRVTT-QA (left) and MSVD-QA (right) benchmarks. The results are reported as percentages (\%), with the \textbf{best} outcomes highlighted in bold.}
\label{table:test-datasets}
\begin{tabular}{lcccccccccccc}
\toprule
\multirow{2}*{\textbf{Method}} & \multicolumn{6}{c}{{\textbf{MSRVTT-QA}}}  & \multicolumn{6}{c}{{\textbf{MSVD-QA}}} \\
\cline{2-13}
\rule{0pt}{9pt} & \textbf{What} & \textbf{Who} & \textbf{How} & \textbf{When} & \textbf{Where} & \textbf{All} & \textbf{What} & \textbf{Who} & \textbf{How} & \textbf{When} & \textbf{Where} & \textbf{All}\\
\midrule
{EVQA (2018) \cite{zeng2017leveraging}}  & 18.9 & 38.7 & 83.5 & 70.5 & 29.2 & 26.4 & 9.7 & 42.4 & \underline{83.8} & 72.4 & 53.6 & 23.3\\
{CoMem (2018) \cite{gao2018motion}}   &23.9 & 42.5 & 74.1 & 69.0& 42.9 & 32.0  & 19.6  & 48.7 & 81.6 & \underline{74.1} & 31.7 & 31.7\\
{HME (2019) \cite{fan2019heterogeneous}} & 26.5 & 43.6 & 82.4 & 76.0 & 28.6 & 33.0 & 22.4 & 50.1 & 73.0 & 70.7 & 42.9 & 33.7\\
{CAN (2019) \cite{yu2019compositional}}  & 26.7 & 43.4 & 83.7 & 75.3 &  35.2 & 33.2 & 21.1 & 47.9 & \textbf{84.1}  & \underline{74.1} & \underline{57.1}  & 32.4\\ 
{STVQA (2019) \cite{jang2019video}}  & 24.5 & 41.2 &78.0 & 76.5 & 34.9  & 30.9 & 18.1 & 50.0 & \underline{83.8} & 72.4 & 28.6 & 31.3\\
{TSN (2019) \cite{yang2019question}} & 27.9 & 46.1 & 84.1 & \underline{77.8} & 37.6 & 35.4 & 25.0 & 51.3 & \underline{83.8} &  \textbf{78.4} & \textbf{59.1} & 36.7\\ 
{HGA (2020) \cite{jiang2020reasoning} }  & 29.2 & 45.7 & 83.5 & 75.2 & 34.0 & 35.5 & 23.5 & 50.4 & 83.0 & 72.4 & 46.4 & 34.7\\
{MHMAN (2020) \cite{yu2020long}} &  28.7 & 47.1 & \underline{85.1} & 77.1 & 35.2 & 35.6 & 23.3 & 50.7 & \textbf{84.1} & 72.4 & 53.6 & 34.6\\
{DualVGR (2021) \cite{wang2021dualvgr} }  & 29.4 & 45.6 & 79.8 & 76.7 & 36.4 & 35.5 & 28.7 & 53.9 & 80.0 & 70.7 & 46.4 & 39.0\\
{VGT (2022) \cite{xiao2022video} }  & - & - & - & - & - & 39.7 & - & - & - & - & - & -\\
{CoVGT (2023) \cite{xiao2023contrastive} }  & - & - & - & - & - & 38.3 & - & - & - & - & - & -\\
\midrule
{ALPRO (2022) \cite{li2022align}}  & \underline{36.0} & \textbf{51.7} & \textbf{85.7} & \textbf{79.6} &  \underline{42.3} & \underline{41.9} & \underline{36.0} & \underline{57.0} & 82.2 & 72.4 & 46.6 & \underline{44.7}\\
{VGT (PT) (2022) \cite{xiao2022video} }  & - & - & - & - & - & 39.7 & - & - & - & - & - & -\\
{MGIN (2023) \cite{wang2023multi}}  & - & - & - & - & - & 38.2 & - & - & - & - & - & 39.7\\
{All-in-one+ (2023) \cite{ko2023open} }  & - & - & - & - & - & 39.5 & - & - & - & - & - & 43.8\\
{CoVGT (PT) (2023) \cite{xiao2023contrastive} }  & - & - & - & - & - & 40.0 & - & - & - & - & - & -\\
\midrule
{HeurVidQA (Ours)}  & \textbf{36.1} & \underline{51.5} & 83.1 & 77.4 &  \textbf{46.2} & \textbf{42.1} & \textbf{36.4} & \textbf{60.8} & 83.5 & 72.4  & 50.0 & \textbf{46.4}\\
\bottomrule
\end{tabular}
\end{table*}

\begin{table*}
\centering
\caption{
Experimental comparative results of the proposed HeurVidQA model and existing SOTA models on SUTD-TrafficQA. Overall accuracy is presented as a percentage (\%). The best outcomes are highlighted in \textbf{bold}. B: Basic understanding, F: Forecasting task, R: Reverse Reasoning, C: Counterfactual inference, I: Introspection, A: Attribution. (C) and (A) refer to prompt addition strategies, while (C) and (C*) denote training prompts with/without adapter heads.
}
\label{table:trafficqa}
\begin{tabular}{l ccccccc}
\toprule
\textbf{Method}  & \textbf{B} & \textbf{F} & \textbf{R} & \textbf{C} & \textbf{I} & \textbf{A} & \textbf{Acc@All}\\
\midrule
CLIP + Template & 31.8 & 36.0 & 29.9 & 41.8 & 22.1 & 33.4 & 32.3 \\
CLIP-Adapter & 35.8 & 32.0 & 35.4 & 42.3 & 33.1 & 32.1 & 34.8 \\
TVQA \cite{lei2018tvqa}  & - & - & - & - & - & - & 35.2 \\
HCRN \cite{fan2019heterogeneous} & - & - & - & - & - & - & 36.5 \\
Eclipse \cite{xu2021sutd} & - & - & - & - & - & - & 37.1 \\
LORA \cite{hu2021lora} & 38.7 & 38.7 & 36.7 & 37.9 & 34.5 & 38.1 & 38.3 \\
Prompt learning (C*) \cite{zhou2022learning} & 40.3 & 33.2 & 41.0 & 46.5 & 34.9 & 38.4 & 39.7 \\
Prompt learning (C) \cite{zhou2022learning} & 42.4 & 32.4 & 45.2 & 55.5 & 40.7 & 43.6 & 42.9 \\
Prompt learning (A) \cite{jia2022visual} & 41.7 & 31.5 & 40.1 & 48.4 & 33.1 & 41.4 & 41.1 \\
Tem-adapter (128 frames) \cite{chen2023tem} & \textbf{46.0} & 36.5 & 44.6 & 55.0 & 34.5 & 47.7 & 46.0 \\
\midrule
HeurVidQA (w/o EAPrompter) & 44.5 & \textbf{40.3} & 51.5 & 48.1 & 36.5 & 48.2 & 45.6\\
HeurVidQA (Ours) & 45.0 & \textbf{40.3} & \textbf{49.2} & \textbf{50.3} & \textbf{41.2} & \textbf{49.5} & \textbf{46.3}\\
\bottomrule
\end{tabular}
\vspace{-10pt}
\end{table*}

\subsubsection{Comparison on NExT-QA}
Table \ref{table:test-std} presents the comparative results on the NExT-QA dataset, encompassing both non-pretrained and pretrained VL models. Notably, we emphasize the baseline method, ALPRO, to demonstrate the improvements achieved through heuristic prompts. The results indicate that pretrained visual-language (VL) models generally surpass non-pretrained models, as evidenced by the performance of VGT and CoVGT compared to their pretrained counterparts, VGT (PT) \cite{xiao2022video} and CoVGT (PT) \cite{xiao2023contrastive}. Our method consistently achieves optimal or near-optimal results across overall performance metrics (Acc@All) and specific question types (Causal, Temporal, and Descriptive). 

It is noteworthy that despite employing ALPRO as the baseline model with identical pre-training weights, HeurVidQA surpasses ALPRO significantly under the same experimental conditions. Specifically, HeurVidQA exhibits a 1.0\% improvement on the validation set and a 1.6\% enhancement on the test set. This performance boost is primarily driven by the strategic incorporation of additional action and entity heuristic information, which significantly enhances the model's fine-grained visual perception capabilities. A focused analysis reveals HeurVidQA's distinct advantages in handling temporal questions, where it outperforms models with larger parameters or more input frames. This superiority stems from the EAPrompter's ability to capture essential action information, which is crucial for answering temporal queries. Notably, this action information, although not explicitly present in the questions or answers, provides precise prompts that significantly improve the model's response accuracy.
Similarly, HeurVidQA excels in descriptive questions, thanks to Entity Prompter's effectiveness in identifying and describing a broad range of entity categories from visual data. By establishing relevant connections between the entities and the questions and answers, the Entity Prompter guides the baseline model in navigating entity-dense video content, thereby enhancing its descriptive question-answering capabilities.
While HeurVidQA may not outperform certain methods in causal reasoning tasks, it is important to recognize that many leading approaches, such as CoVGT (PT) \cite{xiao2023contrastive}, leverage manually crafted complex methodologies, including intricate graph structures designed for logical and causal reasoning. These approaches, combined with doubled input frame settings, naturally excel in causal reasoning tasks. Despite not securing the top spot in this domain, HeurVidQA has made significant progress, particularly in addressing the complexity of causal reasoning content, which requires an integrated analysis of various video elements. The combined efforts of the action and entity prompters within the EAPrompter contribute to notable advancements in our baseline model's performance.

To further validate the effectiveness of our approach in enhancing temporal-spatial reasoning, we conducted experiments on the $ATP_{hard}$ dataset, a subset of the NExT-QA dataset specifically designed to challenge single-frame inference. This subset allows us to assess whether performance improvements are due to genuine enhancements in temporal-spatial reasoning or merely increased model complexity. As shown in Table \ref{table:test-hard}, HeurVidQA surpasses previous models across all question categories, establishing a new state-of-the-art. Remarkably, compared to the baseline model, the HeurVidQA exhibited an approximately 2\% increase in performance on the more challenging dataset, compared to a roughly 1\% enhancement on the original NExT-QA dataset. This differential in performance improvement underscores the capacity of EAPrompter to significantly bolster the answer reasoning capabilities of VideoQA tasks. Importantly, this enhancement is particularly pronounced in datasets that present a higher level of difficulty, highlighting EAPrompter's potential to navigate and address complex VideoQA challenges effectively. Subsequent ablation experiments confirm that Heuristic Prompting significantly contributes to these performance gains.Moreover, we include recently proposed methods \cite{yu2023self, wang2024lstp, yu2024crema} leveraging LLMs, marked in gray due to their large model parameters and extensive pre-training data. Although these methods excel across all metrics, they require significantly more computational resources, with model sizes and data needs far exceeding those of pre-trained vision-language models. Given this disparity, these methods are distinguished in gray.

\subsubsection{Comparison on MSVD-QA and MSRVTT-QA}
To assess the effectiveness of EAPrompter on short video datasets, we conducted experiments on the MSVD-QA and MSRVTT-QA datasets, with results summarized in Table \ref{table:test-datasets}. The findings indicate that HeurVidQA achieves notable improvements on these datasets. While gains on short videos are modest compared to long video datasets, HeurVidQA offers clear advantages. It outperforms VGT and CoVGT by 2.4\% and 3.8\%, respectively, using only half the input frames. This success is largely attributed to EAPrompter’s ability to accurately identify entities and actions, enhancing spatial reasoning. However, the limited duration of the videos in these datasets often leads to a higher prevalence of description and perception questions, which affects overall performance. Despite this, HeurVidQA remains competitive with more complex models that utilize intricate graph structures for logical and causal reasoning. Additionally, while approaches like MGIN focus on multi-level intra-granularity and inter-granularity relations, they may sometimes neglect the relevance of the question text itself. In contrast, HeurVidQA maintains a balanced focus, leading to consistent performance gains.
However, it is important to acknowledge that the limited action and entity information within the MSVD-QA and MSRVTT-QA datasets, combined with the dominance of descriptive questions (e.g., ``What" questions), diminishes the impact of integrating EAPrompter. As a result, the improvements on these datasets are not as significant. Additionally, in the MSRVTT-QA dataset, noise introduced by the Prompter may negatively affect performance, particularly when addressing ``How" questions, further reducing the effectiveness of EAPrompter in these contexts.

\begin{table}
\centering
\caption{Comparison on the ATP-hard subset of the NExT-QA dataset. In addition to the original validation set, we have included a challenging subset identified by ATP that requires video-level understanding for effective evaluation.} 
\label{table:test-hard}
\begin{tabular}{lccc}
\toprule
\textbf{Prompter}  & \textbf{Acc@C} & \textbf{Acc@T} & \textbf{Acc@All}\\
\midrule
ATP \cite{buch2022revisiting} & 19.6 & 22.6 & 20.8 \\
Temporal[ATP] \cite{buch2022revisiting}& 38.4 & 36.5 & 37.6\\
HGA \cite{jiang2020reasoning} & 43.3 &45.3 & 44.1\\
VFC \cite{momeni2023verbs} & 39.9 & 38.3 & 39.3 \\
$\mathrm{SeViT_{FiD}}$ \cite{kim2023semi}& 43.3 & 46.5 & - \\
\midrule
ALPRO \cite{li2022align} & 44.4 & 46.6 & 45.3 \\
HeurVidQA (Ours) & \textbf{46.9} & \textbf{47.8} & \textbf{47.3} \\
\bottomrule
\end{tabular}
\vspace{-10pt}
\end{table}

\subsubsection{Comparison on SUTD-TrafficQA}
To assess HeurVidQA’s generalization across different domains, we evaluated it on the specialized SUTD-TrafficQA dataset, focused on traffic scenarios. Our approach consistently outperformed state-of-the-art methods. Notably, HeurVidQA achieved a 0.3\% improvement over Tem-adapter \cite{chen2023tem}, which uses 126 frames, while HeurVidQA utilized only 16. Significant gains were observed in \emph{Introspection} questions, with a remarkable 6.7\% increase, illustrating the model’s enhanced ability to analyze self-reflective queries within video content. Conversely, the \emph{Counterfactual inference} questions, which involve hypothetical scenarios (e.g., speculating outcomes had the blue car not accelerated), proved more challenging. HeurVidQA trailed the best-performing method by 4.7\%, a limitation likely due to the pre-trained data that lacks specialized knowledge in traffic scenarios. While HeurVidQA exhibits commendable inferential capabilities across various domains, its performance in complex reverse reasoning tasks remains constrained. Nevertheless, the model showed a significant improvement over the baseline across all question types, underscoring EAPrompter’s effectiveness in enhancing reasoning abilities, especially within traffic video QA tasks.

\begin{table}
\centering
\caption{Comparison of different prompt strategies. ``w/ Prompter Gate" refers to the integration of a gating mechanism that automatically adjusts the weight distribution between entity and action prompts. The ratios in parentheses indicate the relative proportion of entity-to-action prompts within the overall heuristic pool.}
\label{table:test-prompt}
\begin{tabular}{lcccc}
\toprule
Prompter  & \textbf{Acc@C} & \textbf{Acc@T} & \textbf{Acc@D} & \textbf{Acc@All}\\
\midrule
w/o EAPrompter & 57.61 & 57.75 & 71.69 & 59.85\\
\midrule
w/ EP (1:0) & \underline{58.42} & 58.31 & 71.81 & 60.47\\
w/ AP (0:1) & 57.81 & \textbf{58.56} & \textbf{73.75} & \underline{60.53}\\
\midrule
w/ EAPrompter (1:1) & 57.19 & 58.31 & 71.3 & 59.75\\
w/ EAPrompter Gate & \textbf{58.84} & \underline{58.37} & \underline{72.97} & \textbf{60.89}\\
\bottomrule
\end{tabular}
\vspace{-10pt}
\end{table}

\begin{figure*}
\centering
  \subfigure[Acc@C]{\includegraphics[width=0.23\textwidth]{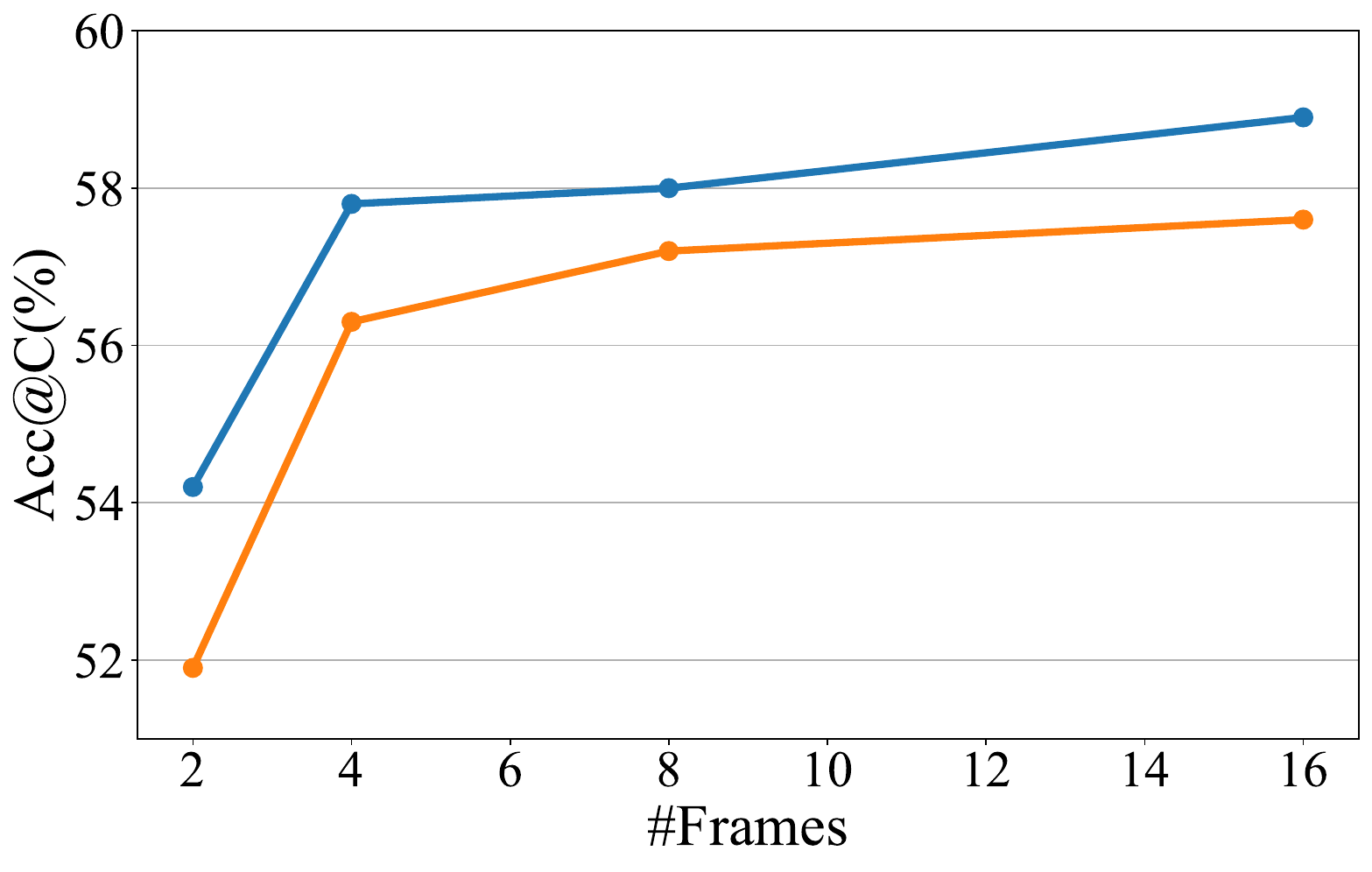}\label{fig: sub_figure1-1}}\hspace{0.5cm}
  \subfigure[Acc@D]{\includegraphics[width=0.22\textwidth]{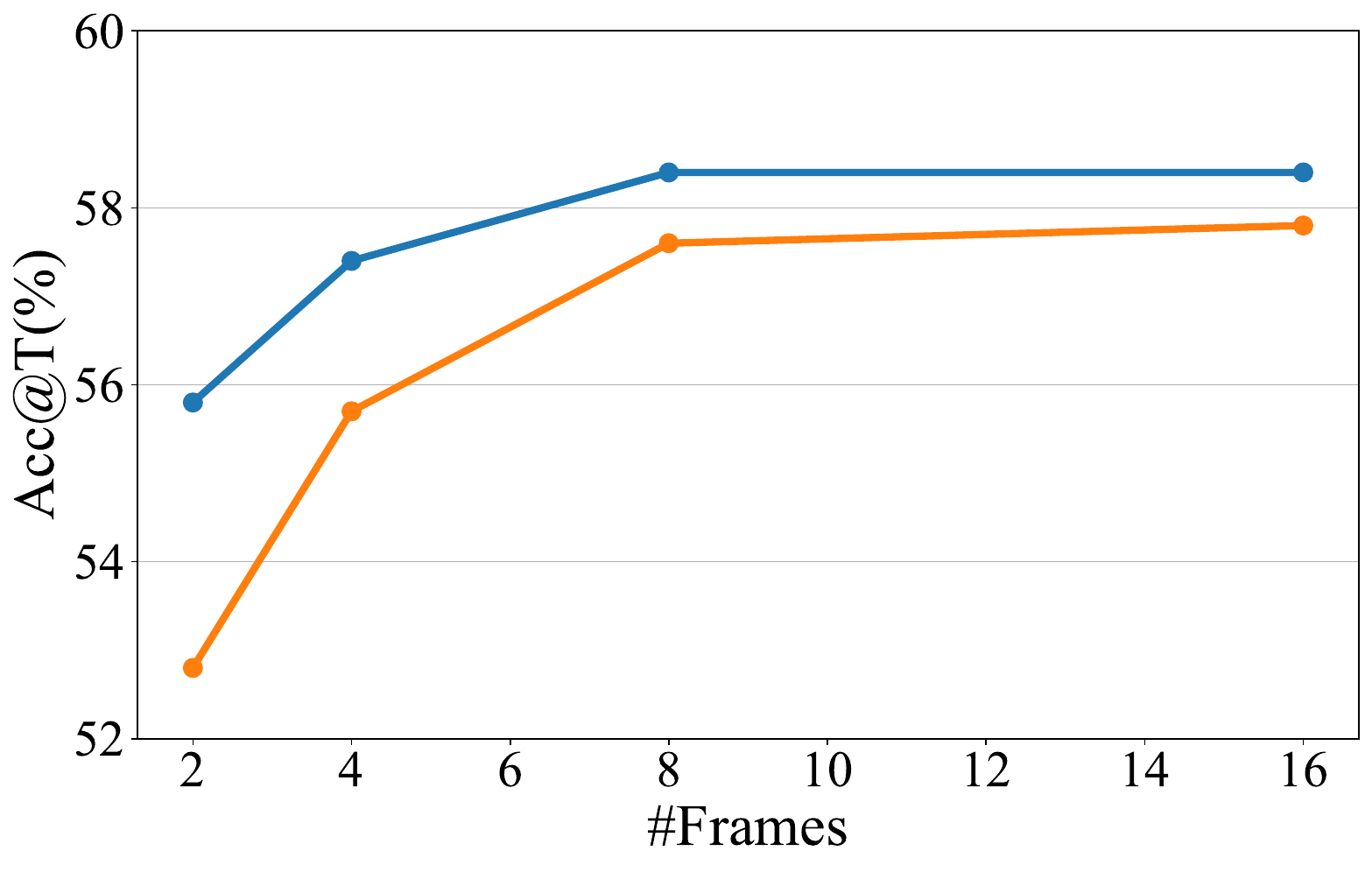}\label{fig: sub_figure1-2}}\hspace{0.5cm}
  \subfigure[Acc@T]{\includegraphics[width=0.22\textwidth]{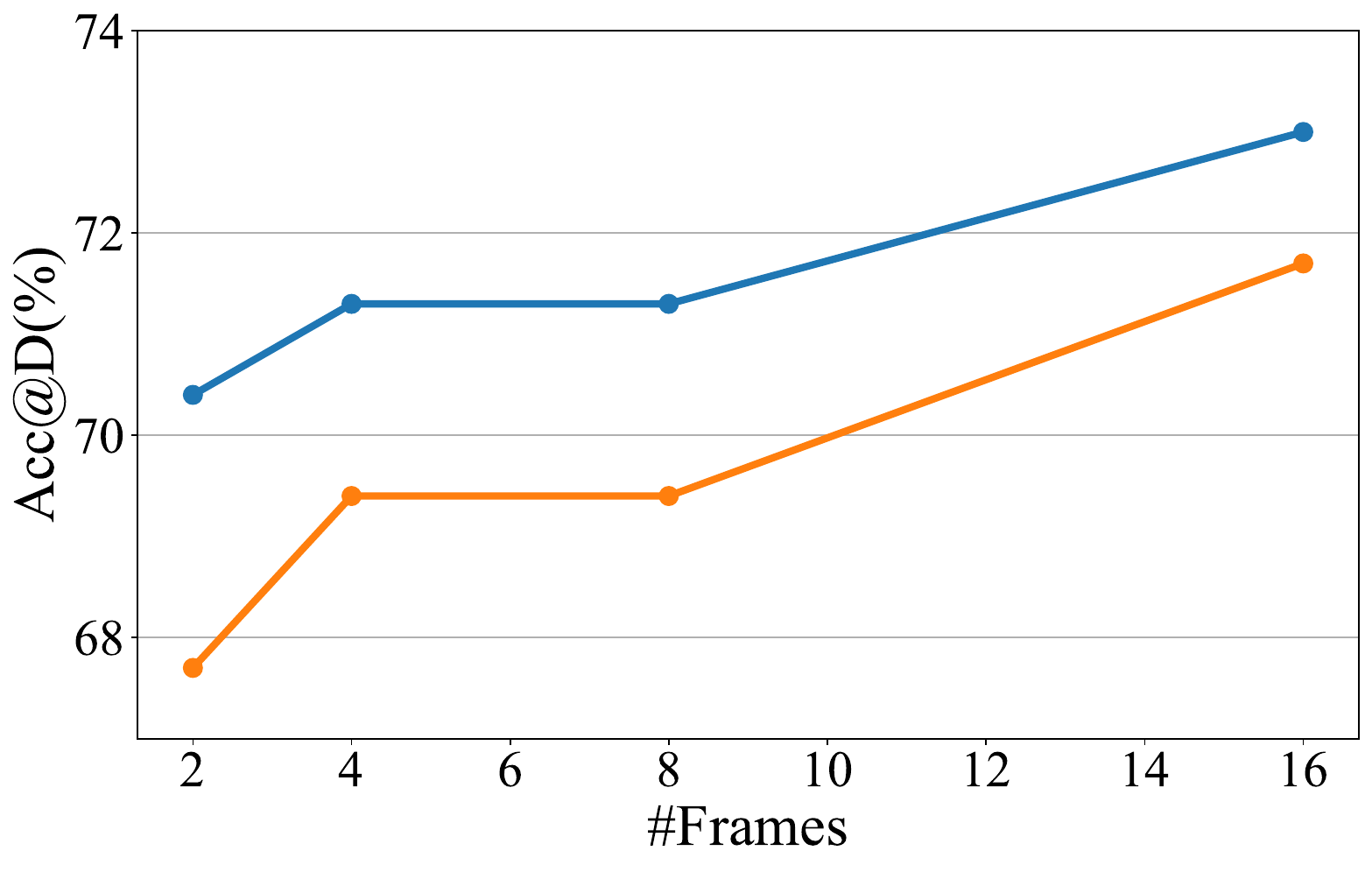}\label{fig: sub_figure1-3}}\hspace{0.5cm}
  \subfigure[Acc@All]{\includegraphics[width=0.22\textwidth]{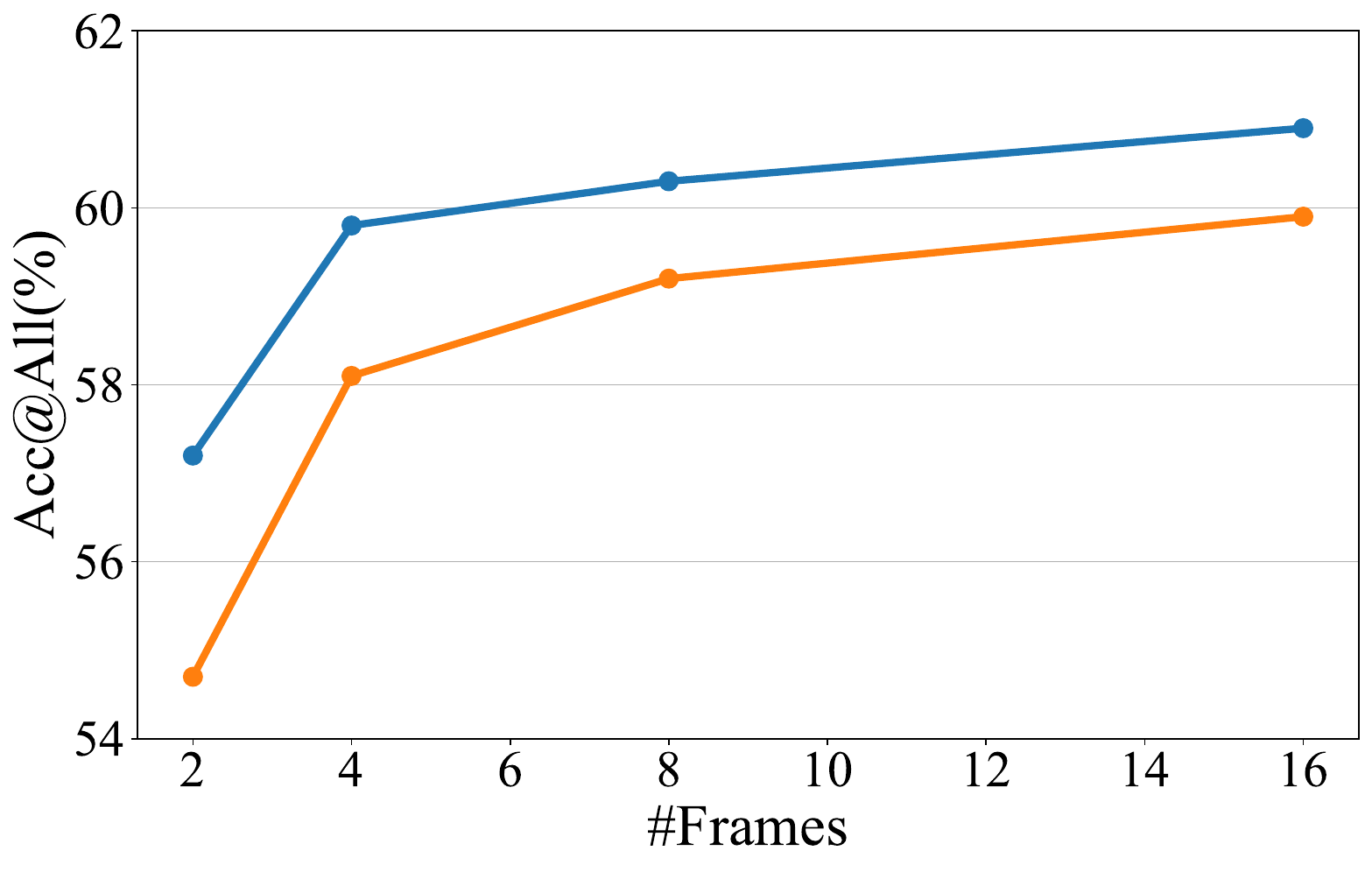}\label{fig: sub_figure1-4}}\hspace{0.5cm}
\caption{
Variations in test accuracy with different video frame sampling strategies across question types: (a) Causal, (b) Descriptive, (c) Temporal, and (d) overall performance within the NExT-QA dataset. Sampling frames range from 2 to 16, showing consistent performance gains as frame count increases. The \textcolor[RGB]{29, 108, 171}{blue line} represents HeurVidQA’s performance, while the \textcolor[RGB]{255, 139, 57}{orange line} shows baseline ALPRO’s performance. Notably, causal and descriptive questions improve significantly with 4 frames, while temporal questions benefit more at 16 frames.
 }
\label{fig:video_frame}
\vspace{-10pt}
\end{figure*}

\subsection{Ablation Study}
We conducted several ablation experiments on the NExT-QA dataset to explore the effects of different factors, including heuristic ratios, sparsity versus density, heuristic quantities, and prompt templates. The goal was to understand how each component influences the overall performance of HeurVidQA. Aside from these specific parameters, all other experimental settings remained consistent with those defined earlier. The findings from these ablation studies provide deeper insights into the critical elements that contribute to the effectiveness of our proposed framework.

\begin{figure}
\centering
  \subfigure[Acc@C]{\includegraphics[width=0.23\textwidth]{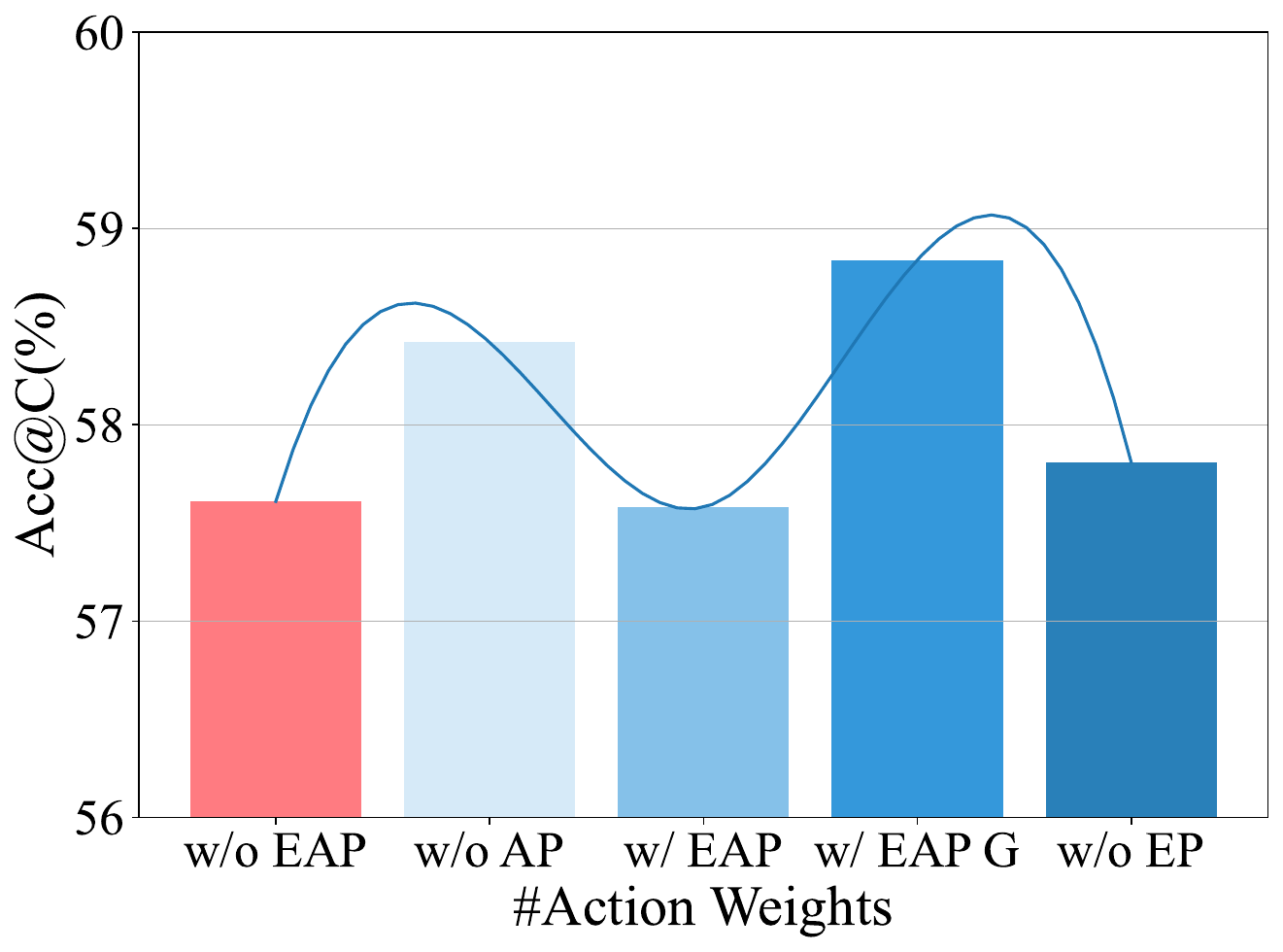}\label{fig: sub_figure2-1}}\hspace{0.5cm}
  \subfigure[Acc@T]{\includegraphics[width=0.22\textwidth]{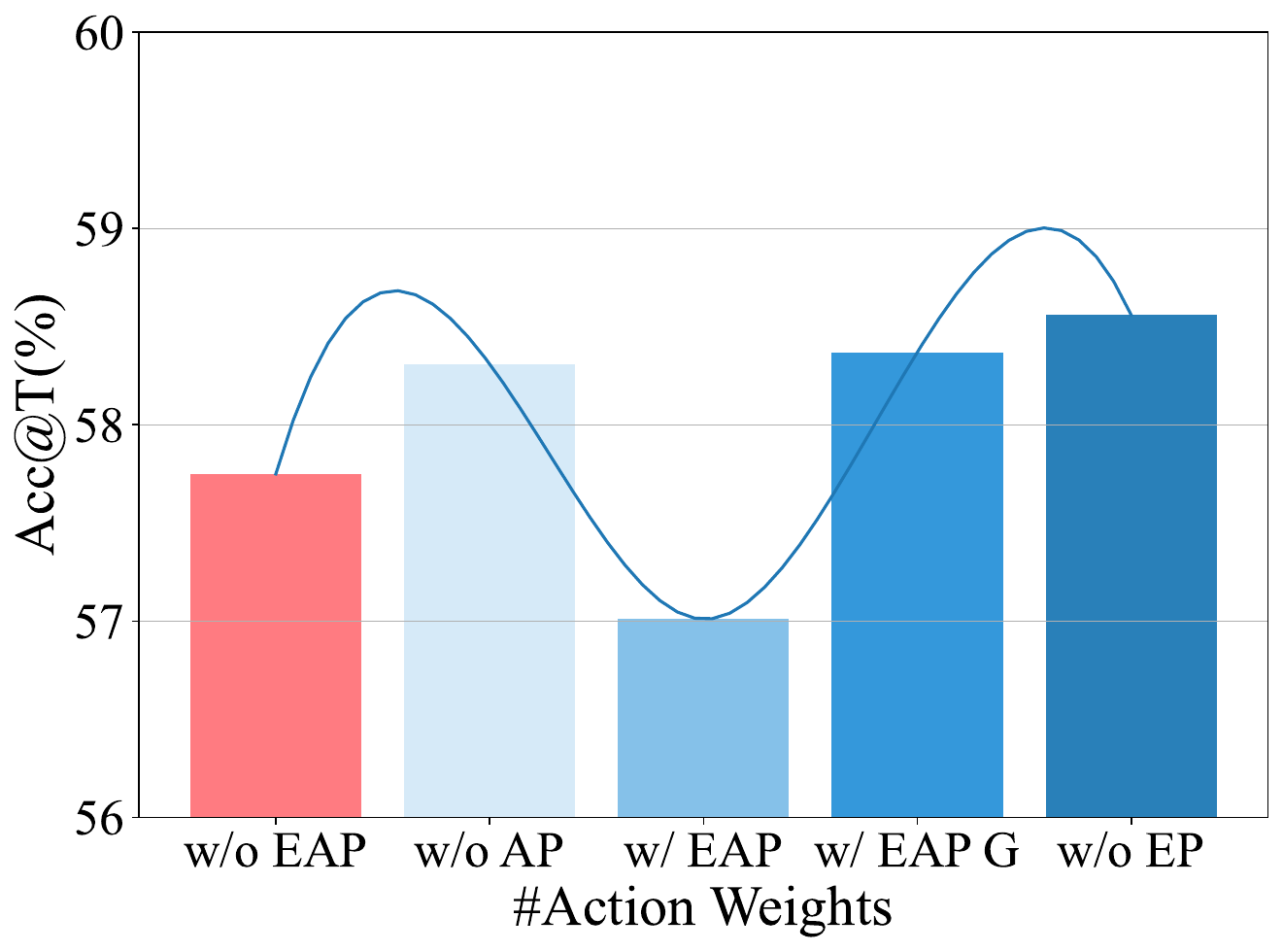}\label{fig: sub_figure2-2}}\hspace{0.5cm}
  \subfigure[Acc@D]{\includegraphics[width=0.22\textwidth]{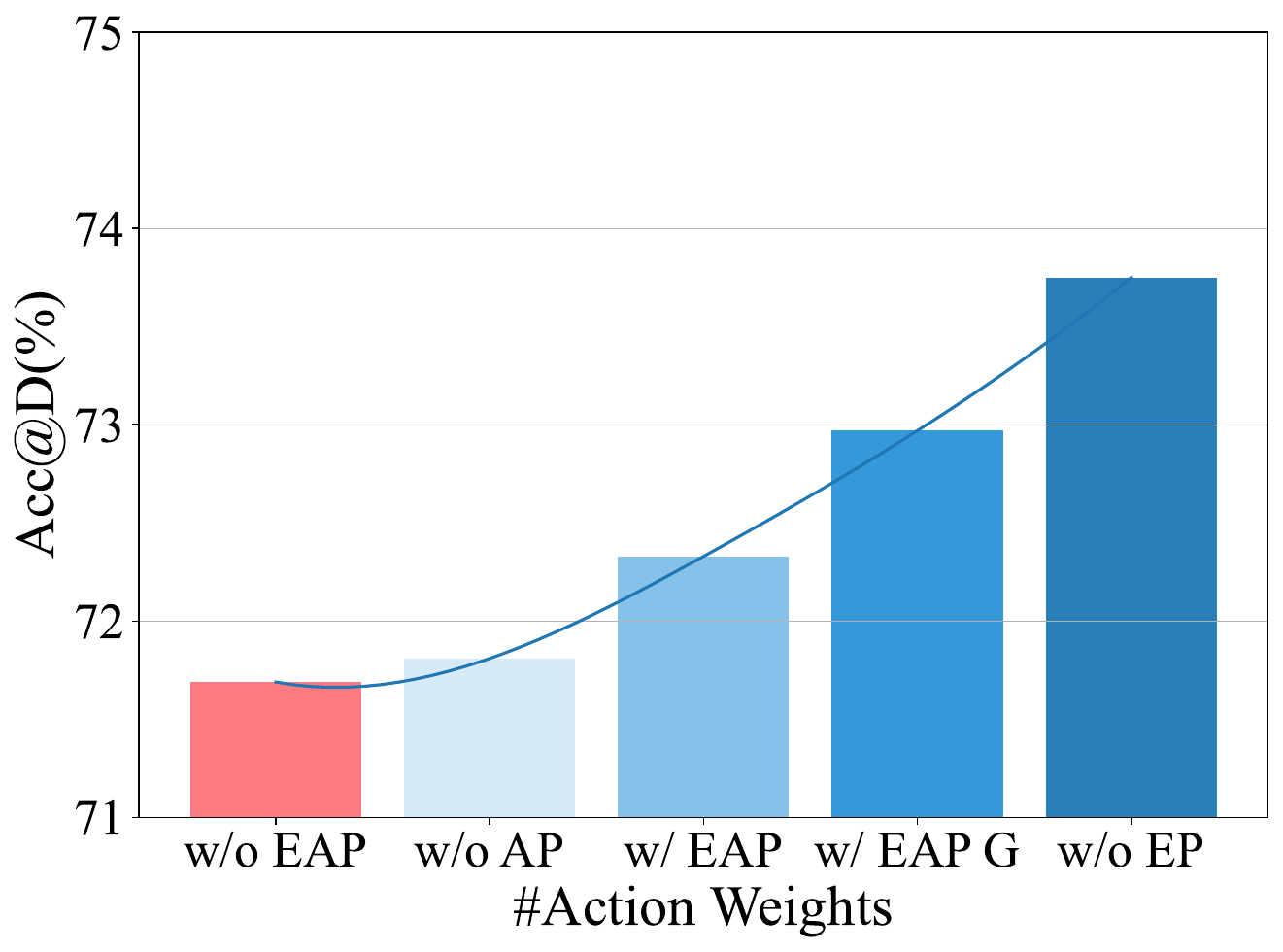}\label{fig: sub_figure2-3}}\hspace{0.5cm}
  \subfigure[Acc@All]{\includegraphics[width=0.22\textwidth]{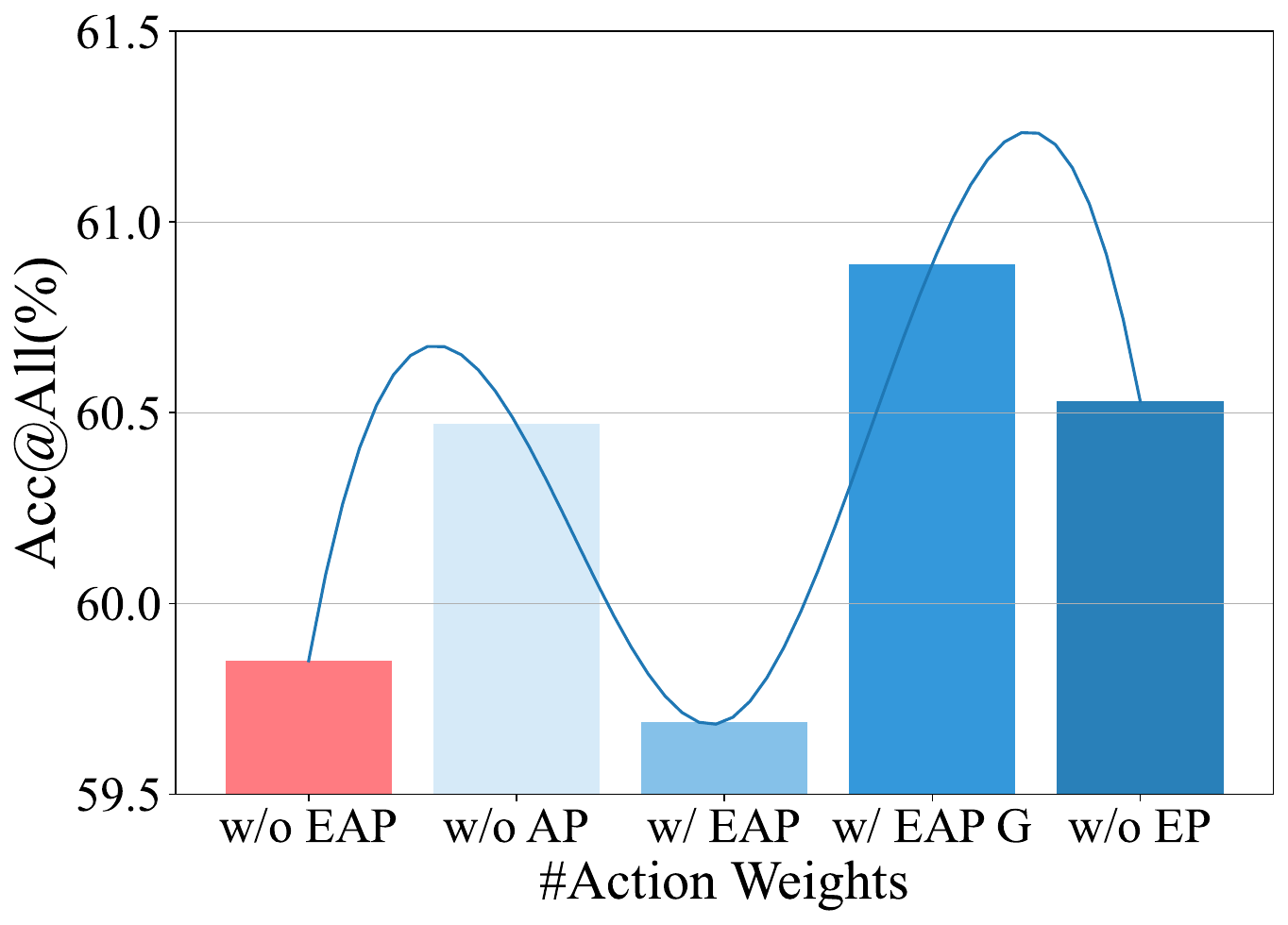}\label{fig: sub_figure2-4}}\hspace{0.5cm}
\caption{
Analysis of the impact of different heuristic prompt components. The visualization shows model performance with varying proportions of action prompts. The red column indicates the baseline model without heuristic augmentation, while deeper blue columns represent higher action prompt ratios within the heuristic set. The curve illustrates the trend in model effectiveness as action prompts increase. EAP: EAPrompter, AP: Action Prompter, EP: Entity Prompter, EAP G: EAPrompter Gate.
}
\label{fig:components}
\vspace{-10pt}
\end{figure}

\subsubsection{Heuristic Ratios}
To evaluate the impact of Heuristic Prompting, we conducted experiments by not only adding or removing the prompters but also varying the ratio between action and entity heuristics. We compared fixed heuristic ratios against a dynamic ratio adjusted by a gating mechanism. 
Table \ref{table:test-prompt} details the results, showing a clear performance boost over the baseline model. Notably, the degree of improvement is influenced by the ratio of heuristic prompts used.
Introducing either the Action Prompter or Entity Prompter significantly enhanced the model's performance. However, using both prompts together yielded mixed results depending on their relative weighting.
To achieve an optimal ratio, we incorporated a gating mechanism, enabling the model to dynamically calculate the appropriate proportion of each prompt based on the characteristics of the current training sample. This approach consistently improved performance, as visualized in Figure \ref{fig:components}, where we gradually increased the proportion of action prompts.
Our analysis showed that when the model lacked action prompts, adding entity prompts only marginally improved performance, as the baseline model already incorporated substantial entity information during pre-training. However, equalizing the action and entity prompts initially introduced noise, leading to performance fluctuations. Further increasing the action prompt proportion significantly enhanced performance, indicating that both action and entity prompts are essential for the model, despite substantial pre-training emphasis on entity information.

\begin{table}
\vspace{-5pt}
\centering
\caption{Detailed results on the NExT-QA benchmark with varying numbers of frames.}
\label{table:test-frames}
\begin{tabular}{lcccc}
\toprule
\textbf{\#frms} & \textbf{Acc@C} & \textbf{Acc@T} & \textbf{Acc@D} & \textbf{Acc@All} \\
\midrule
\multicolumn{4}{c}{\textit{\begin{footnotesize}{Baseline (ALPRO) frame ablation experiment}\end{footnotesize}}} \\
1 & {51.9} & {52.8} & {67.7} & {54.7} \\
4 & {56.3} & {55.7} & {69.4} & {58.1} \\
8 & {57.2} & {57.6} & {69.4} & {59.2} \\
16 & 57.6 & \underline{57.8} & \underline{71.7} & 59.9 \\
\midrule
\multicolumn{4}{c}{\textit{\begin{footnotesize}{HeurVidQA (ours) frame ablation experiment}\end{footnotesize}}} \\
1 & 54.2 & 55.8 & 70.4 & 57.2\\
4 & 57.8 & 57.4 & 71.3 & 59.8\\
8 & \underline{58.0} & \textbf{58.4} & 71.3 & \underline{60.3}\\
16 & \textbf{58.8} & \textbf{58.4} & \textbf{73.0} & \textbf{60.9}\\
\bottomrule
\vspace{-15pt}
\end{tabular}
\end{table}

\subsubsection{Sparse Vs. Dense}
Figure \ref{fig:video_frame} presents the ablation results for varying numbers of sampled frames per video, with candidate values of $N\in\{1, 4, 8, 16\}$. We compared HeurVidQA against the baseline under identical conditions. The blue line delineates the performance metrics of HeurVidQA, whereas the orange line illustrates the performance of the baseline model.
Upon incrementally augmenting the number of input video frames, both the baseline model and HeurVidQA exhibited similar trends in performance. However, HeurVidQA consistently outperformed the baseline model across all frame configurations, demonstrating the comprehensive enhancement imparted by the EAPrompter.
Table \ref{table:test-frames} provides detailed experimental results, clearly showing the superiority of our method over the baseline, even with fewer sampled frames. The influence of additional frames varies by question type: causal questions benefit the most (around a 4.7\% increase), while temporal and descriptive questions see similar improvements (about 2.6\% each). Temporal questions likely demand more frames, yet merely adding frames does not suffice for major gains. Descriptive questions, less dependent on temporal data, achieve 70.4\% effectiveness with just one frame but still benefit from additional frames.

\subsubsection{Impact of actions and entities candidate quantities}
We conducted a comprehensive ablation study to assess the impact of actions and entities candidate quantities on model performance, with results detailed in Table \ref{table:heuristic-numbers}. The first row showcases the baseline model's performance using four frames without incorporating heuristics. We then systematically increased the number of verbs and nouns from 400 to 2000. The findings suggest that a balanced quantity of heuristics significantly improves visual-text alignment. However, an excessive number of heuristics introduces low-frequency elements and noise, which can degrade model performance.

\subsubsection{Impact of prompt templates}
The design of prompt templates is critical to the effectiveness of the EAPrompter, as it directly influences the extraction of entity and action heuristics from video content. To explore this, we crafted 10 distinct templates for various heuristic types, as shown in Table \ref{table:prompt-templates}. We then conducted an ablation study using both the long video dataset NExT-QA and the short video dataset MSVD-QA.
Interestingly, the impact of prompt templates varied across datasets, as illustrated in Table \ref{table:prompt-templates-experiment}. For the NExT-QA dataset, which features long videos with abundant actions and entities, using complex prompt templates led to a slight performance decline. This suggests that more confident heuristics from refined templates might inadvertently distract the model, leading to errors in reasoning and answer selection. Conversely, the MSVD-QA dataset, which focuses on descriptive and perceptual queries, benefited significantly from the optimized templates. This enhancement indicates that refined templates help provide richer visual heuristics, thereby improving the model's performance in perception-based tasks.

\begin{table}
\vspace{-5pt}
\centering
\caption{
Results vary with different quantities of candidate actions and entities.}
\label{table:heuristic-numbers}
\begin{tabular}{lcccc}
\toprule
\#count  & Acc@C & Acc@T & Acc@D & Acc@All\\
\midrule
\makecell[c] {-} & 57.6& 57.8 & 71.7 & 59.9 \\ 
\midrule
\makecell[c] {400}  & 58.1 & 58.4 & 73.5 & 60.6\\
\makecell[c] {600} & 58.4 & 59 & 73.1 & 60.9\\
\makecell[c] {1000} & 58.2 & \textbf{59.4} & \textbf{73.6} & \textbf{61}\\
\makecell[c] {1600} & \textbf{58.7} & 58.4 & 72.7 & 60.8\\
\makecell[c] {2000} & 57.3 & 58.3 & 71.8 & 60.2 \\ 
\bottomrule
\end{tabular}
\end{table}

\begin{table}
\centering
\caption{Prompt templates for instantiating text prompts.}
\label{table:prompt-templates}
\begin{tabular}{l|l}
\toprule
\textbf{Template Type} & \textbf{Prompt templates}\\
\midrule
\multirow{10}*{\textbf{Entities}} & A video of a \{\}.\\
 & A video of the entity \{\}. \\
 & A video contains the entity of \{\}. \\
 & A shooting of a \{\}. \\
 & A shooting of the \{\}. \\
 & A shooting contains the entity of \{\}. \\
 & A video footage of a \{\}. \\
 & A video footage of the \{\}. \\
 & A footage contains the entity of \{\}. \\
 & A video recording about the entity of \{\}. \\
 \midrule
\multirow{10}*{\textbf{Actions}} & A video contains the action of \{\}. \\
 & A video about the action of \{\}. \\
 & A video recording about the action of \{\}. \\
 & A video shooting of the action \{\}. \\
 & A video of action \{\} being performed. \\
 & A footage of the action of \{\}. \\
 & A shooting of the action \{\}. \\
 & A shooting of \{\} in action. \\
 & A clip of \{\} in action. \\
 & A clip contains the action of \{\}.\\
\bottomrule
\end{tabular}
\vspace{-10pt}
\end{table}

\begin{table}
\centering
\caption{
Experimental results using different types of prompt templates. ``Simple Temp." refers to simple and identical prompt templates, while ``Complex Temp." involves using templates specifically tailored for entities and actions
}
\label{table:prompt-templates-experiment}
\begin{tabular}{l|cc}
\toprule
 & \textbf{MSVD-QA} & \textbf{NExT-QA}\\
\midrule
\textbf{Simple Temp.} & 45.3 & \textbf{60.9} \\
\textbf{Complex Temp.} & \textbf{46.6} & 60.2 \\
\bottomrule
\end{tabular}
\vspace{-10pt}
\end{table}
\subsection{Complexity analysis}
\begin{figure}
\begin{center}
\includegraphics[width=0.5\textwidth]{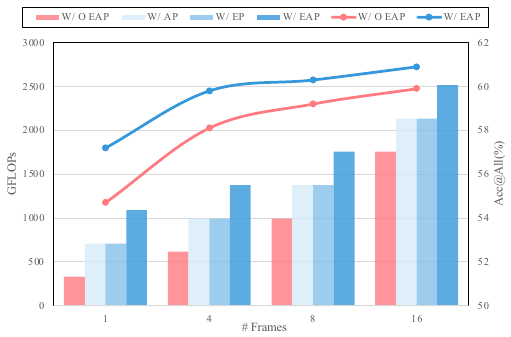}
\caption{The time complexity of models across various configurations and their corresponding performance outcomes. The line chart illustrates the overall performance of HeurVidQA and the comparative models on the NExT-QA dataset. The bar chart shows the GFLOPs required for training a single sample.
}
\label{fig:time_complexity}
\vspace{-20pt}
\end{center}
\end{figure}

We conducted experiments on the model's complexity. Detailed results are provided in Tables \ref{table:space_complexity} and \ref{table:time_complexity}. Table \ref{table:space_complexity} reveals that our model uses less than half the trainable parameters compared to popular contrastive models like JustAsk, ATP, and VGT while achieving superior performance. Notably, leveraging EAPrompter does not increase the number of trainable parameters but introduces additional frozen parameters specifically for generating heuristic information. This increment is justified by the significant performance gains observed. Moreover, the unified design of the EP and AP underscores their importance, as removing either component impacts performance without reducing the total parameter count. This highlights EAPrompter's effectiveness in enhancing VideoQA tasks without substantially increasing model complexity.
In Table \ref{table:time_complexity}, we thoroughly analyze the time complexity of the HeurVidQA model across various configurations, using the Giga Floating Point Operations per Second (GFLOPs) metric to quantify computational demand for training and inference on a single sample. Moreover, our model demonstrates a GFLOP advantage over SeViLA, emphasizing its efficiency. To provide a clearer understanding, Figure \ref{fig:time_complexity} presents visualizations, contrasting HeurVidQA with the baseline model's performance on the NExT-QA dataset. The results indicate that while the integration of EAPrompter marginally increases computational overhead as the number of input frames rises, this increase diminishes as more frames are added. Notably, during inference, HeurVidQA's GFLOPs align closely with the baseline, confirming that EAPrompter does not add to the computational burden in this phase. Thus, the enhancements provided by EAPrompter in HeurVidQA strike a balance between performance gains and computational efficiency, without significantly increasing the model's overall complexity.

\begin{table}
\centering
\caption{
Comparative analysis of model parameters between the baseline and our proposed HeurVidQA model. The table also highlights the performance comparison between our model and the baseline under identical configuration.}
\centering
\label{table:space_complexity}
\begin{tabular}{lcccc}
\toprule
\textbf{Method} & \textbf{\makecell [c]{Trainable \\ param}} & \textbf{\makecell [c]{Total \\ param}} & \textbf{Acc@All} \\
\midrule
{JustAsk} \cite{yang2021just}  & 600M & 600M & 45.3\\
{ATP} \cite{buch2022revisiting} & - & 428M & 54.3\\
{VGT} \cite{xiao2022video} & 511M & 511M & 55.7\\
{PAXION} \cite{wang2024paxion} & 8.2M & 482M & 56.9\\
\midrule
{HeurVidQA w/o EAP} & 237M & 237M & 59.9 \\ 
{HeurVidQA w/ AP} & 237M & 469M & 60.5 \\ 
{HeurVidQA w/ EP} & 237M & 469M & 60.4 \\ 
{HeurVidQA (ours)} & 237M & 469M & \textbf{60.9} \\ 
\bottomrule
\end{tabular}
\end{table}

\begin{figure*}
\begin{center}
\includegraphics[width=1\textwidth]{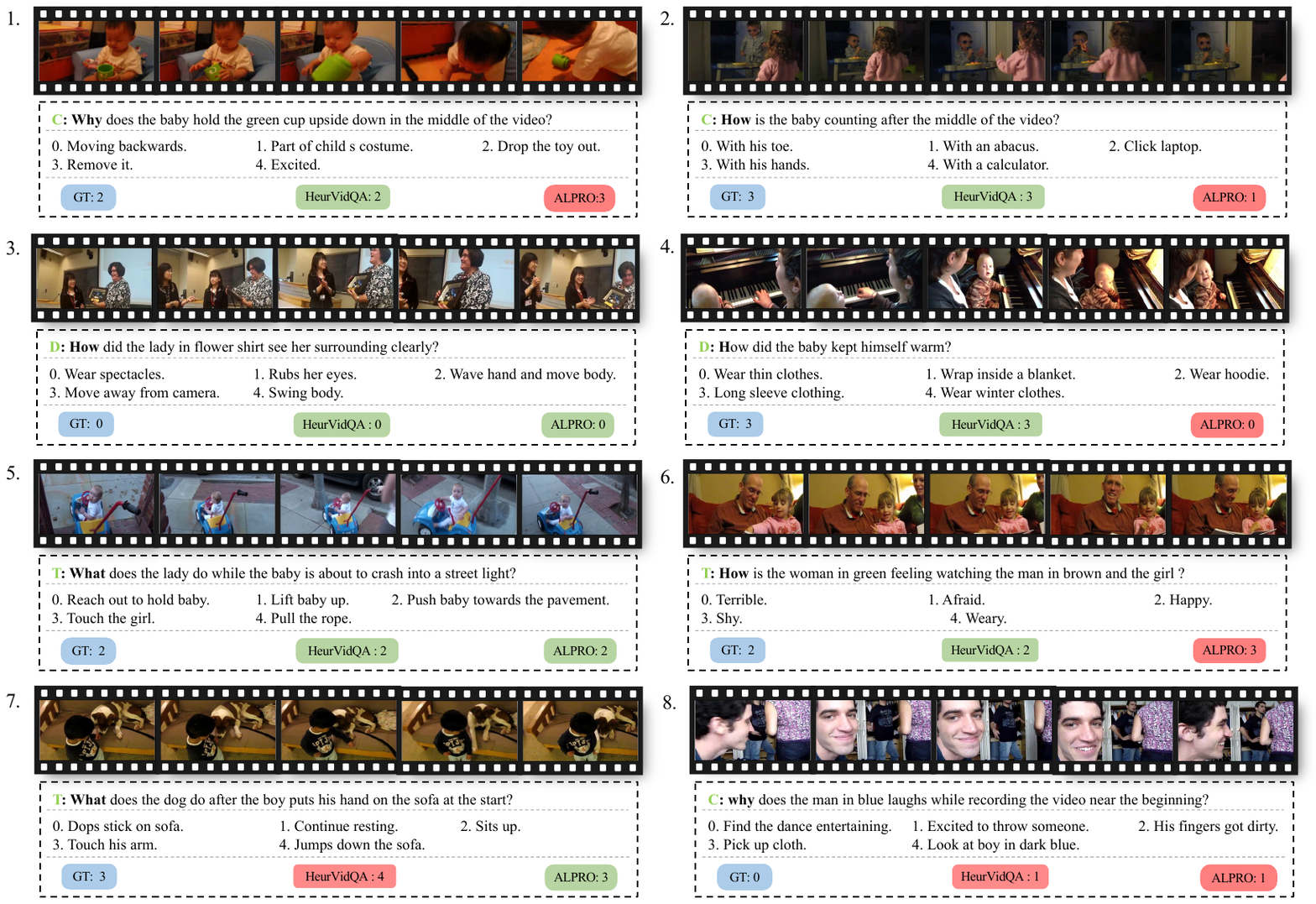}
\caption{Visualization of the question-answering results generated by the proposed HeurVidQA model on the NExT-QA dataset. The green letter at the top of each example indicates the question type: C for Causal, T for Temporal, and D for Descriptive. The ground truth answer, along with the predictions from HeurVidQA and the baseline ALPRO model, is shown at the bottom of each example. Correct answers are highlighted in green, while incorrect answers are marked in red.}
\label{fig:visiualization}
\end{center}
\vspace{-10pt}
\end{figure*}

\begin{table}
\centering
\caption{
Comparative analysis of model Giga Floating Point Operations per Second (GFLOPs) metrics across various configurations. The overall performance indicators for the NExT-QA dataset are included for a consolidated comparison.
}
\centering
\label{table:time_complexity}
\begin{tabular}{lccc}
\toprule
\textbf{Method} & \textbf{\makecell [c]{GFLOPs}}  & \textbf{Acc@All} \\
\midrule
{SeViLA \cite{yu2023self}} & 51181 & 73.4 \\ 
{ALPRO \cite{li2022align}} & 439 & 59.9 \\ 
\midrule
{HeurVidQA w/ AP} & 534  & 60.5 \\ 
{HeurVidQA w/ EP} & 534 & 60.4 \\ 
{HeurVidQA (ours)} & 629 & \textbf{60.9} \\ 
\bottomrule
\end{tabular}
\end{table}

\subsection{Qualitative Analysis}
Figure \ref{fig:visiualization} showcases the prediction outcomes generated by HeurVidQA on various question categories within the NExT-QA dataset. Each question is prefixed with a letter indicating its type, C for Causal, T for Temporal, and D for Descriptive.
To comprehensively evaluate our approach, we analyze HeurVidQA with the comparative model ALPRO. Three key observations emerged from this analysis:
(1) Both HeurVidQA and ALPRO effectively address questions with extended temporal scopes, such as comprehending the lady's behavior in Example 5.
(2) HeurVidQA surpasses ALPRO in localizing smaller visual objects and reasoning about challenging questions. For instance, in Example 4, where the video captures a confined spatial and temporal presence of women, HeurVidQA's enhanced performance is attributed to its heuristic prompts, which enable it to detect and identify diverse entities and actions, leading to improved sensitivity in temporal reasoning questions.
(3) HeurVidQA outperforms ALPRO on more complex long-term questions. This advantage stems from its augmented language modeling capabilities, facilitated by heuristic prompts, allowing the model to manage multiple actions and physical information, ultimately contributing to superior performance.
(4) While HeurVidQA demonstrates excellent performance in addressing descriptive questions, its efficacy is constrained when tackling intricate causal and temporal scenarios. This limitation is primarily attributed to the model's reliance on explicitly derived action entities from video content as primary sources of heuristics. Such an approach, while effective for straightforward descriptive tasks, does not inherently equip HeurVidQA with the nuanced inferential capabilities requisite for dissecting and addressing more complex reasoning challenges. Consequently, the model's performance in managing scenarios that demand high-level complex reasoning exhibits certain boundaries.

\vspace{-10pt}
\section{Conclusions and Future Directions }
This paper presents the HeurVidQA model, which leverages principles of prompt learning and engineering to enhance the capabilities of pre-trained visual language models in VideoQA tasks. By integrating heuristic prompts that target both action information across temporal frames and entity information across spatial regions, we introduce two novel loss functions: Prompt Action Modeling and Prompt Entity Modeling. Additionally, we propose a dynamic gating mechanism to maintain a balanced emphasis between action and entity prompts,optimizing the model’s reasoning and inference capabilities.

While HeurVidQA demonstrates substantial improvements in VideoQA tasks, it is not without limitations. A key limitation lies in its reliance on domain-specific heuristic prompts, which, while effective, may restrict its generalization across significantly different domains. Additionally, the model’s performance in handling highly abstract or counterfactual reasoning scenarios remains limited, as it predominantly leverages explicit action-entity heuristics derived from video content. To address these limitations, future work could explore adaptive prompt learning mechanisms that dynamically adjust to diverse domains without requiring manual heuristic definitions. Incorporating more advanced reasoning techniques, such as causal inference and counterfactual analysis, could further enhance the model’s robustness in complex scenarios. Additionally, integrating knowledge graphs or external world knowledge may provide the model with a richer contextual understanding, enabling it to handle a broader range of VideoQA challenges. Finally, reducing the model’s dependency on large-scale pretraining data could improve its scalability and applicability to resource-constrained environments.

\bibliographystyle{IEEEtran}
\bibliography{TCSVT2024}

\vspace{-4pt}
\end{document}